\documentclass[acmsmall,nonacm]{acmart}

\usepackage{lipsum}
\usepackage{amsfonts}
\usepackage{graphicx}
\usepackage{epstopdf}
\usepackage{algorithmic}
\usepackage{bbm}
\usepackage{bm}
\usepackage{tikz}
\usepackage{booktabs} %
\usepackage{siunitx}
\usepackage{multirow}
\usepackage{cleveref}
\usepackage{natbib}
\usepackage[frozencache]{minted}
\setminted{fontsize=\scriptsize}

\usepackage{amsopn}

\newcommand{\mat}[1]{\bm{{#1}}}
\renewcommand{\vec}[1]{\bm{{#1}}}

\newcommand{\set}[1]{{\mathcal{#1}}}

\newcommand{\grad}{\nabla}

\DeclareMathOperator{\mask}{mask}
\DeclareMathOperator{\vectorize}{vec}

\newcommand{\gradfn}[2]{\nabla_{{#1}}({#2})} %
\newcommand{\jac}[2]{J_{{#1}}({#2})} %

\AtBeginDocument{%
  }

\setcopyright{acmcopyright}
\copyrightyear{2018}
\acmYear{2018}
\acmDOI{XXXXXXX.XXXXXXX}

\acmConference[qwerty]{asdf}{June 03--05,
  2018}{Woodstock, NY}
\acmPrice{15.00}
\acmISBN{978-1-4503-XXXX-X/18/06}  %

\acmJournal{JACM}
\acmVolume{37}
\acmNumber{4}
\acmArticle{111}
\acmMonth{8}

\definecolor{tab-blue}{HTML}{1f77b4}
\definecolor{tab-orange}{HTML}{ff7f0e}
\definecolor{tab-green}{HTML}{2ca02c}
\definecolor{tab-red}{HTML}{d62728}
\definecolor{tab-purple}{HTML}{9467bd}
\definecolor{tab-brown}{HTML}{8c564b}
\definecolor{tab-pink}{HTML}{e377c2}
\definecolor{tab-gray}{HTML}{7f7f7f}
\definecolor{tab-olive}{HTML}{bcbd22}
\definecolor{tab-cyan}{HTML}{17becf}
\usetikzlibrary{arrows.meta}
\usetikzlibrary{tikzmark}
\usetikzlibrary{calc}

\begin{document}

\newcommand{\thetitle}{Optimized Sparse Matrix Operations for Reverse Mode Automatic Differentiation}
\newcommand{\thepdfauthors}{Nytko, Taghibakhshi, Zaman, MacLachlan, Olson, and West, Matthew}

\title{\thetitle}

\author{Nicolas Nytko}
\email{nnytko2@illinois.edu}
\affiliation{%
  \institution{University of Illinois at Urbana-Champaign}
  \department{Department of Computer Science}
  \streetaddress{201 N. Goodwin Ave.}
  \city{Urbana}
  \state{IL}
  \country{USA}
  \postcode{61801-2302} %
}

\author{Ali Taghibakhshi}
\email{alit2@illinois.edu}
\affiliation{%
  \institution{NVIDIA Corporation}
  \city{Santa Clara}
  \state{CA}
  \country{USA}
}\additionalaffiliation{%
  \institution{University of Illinois at Urbana-Champaign}
  \department{Department of Mechanical Science and Engineering}
  \streetaddress{1206 N. Green St.}
  \city{Urbana}
  \state{IL}
  \country{USA}
  \postcode{61801-2302} %
}

\author{Tareq Uz Zaman}
\email{tzaman@mun.ca}
\affiliation{%
  \institution{Memorial University of Newfoundland}
  \department{Department of Process Engineering}
  \city{St.\ John's}
  \state{NL}
  \country{Canada}
}

\author{Scott MacLachlan}
\email{smaclachlan@mun.ca}
\affiliation{%
  \institution{Memorial University of Newfoundland}
  \department{Department of Mathematics and Statistics}
  \city{St.\ John's}
  \state{NL}
  \country{Canada}
}

\author{Luke N. Olson}
\email{lukeo@illinois.edu}
\affiliation{%
  \institution{University of Illinois at Urbana-Champaign}
  \department{Department of Computer Science}
  \streetaddress{201 N. Goodwin Ave.}
  \city{Urbana}
  \state{IL}
  \country{USA}
  \postcode{61801-2302} %
}

\author{Matthew West}
\email{mwest@illinois.edu}
\affiliation{%
  \institution{University of Illinois at Urbana-Champaign}
  \department{Department of Mechanical Science and Engineering}
  \streetaddress{1206 N. Green St.}
  \city{Urbana}
  \state{IL}
  \country{USA}
  \postcode{61801-2302} %
}

\renewcommand{\shortauthors}{Nytko et al.}

\begin{abstract}
  Sparse matrices are ubiquitous in computational science, enabling significant reductions in both compute time and memory overhead for problems with local connectivity.  Simultaneously, recent work in automatic differentiation has lead to explosive growth in the adoption of accessible frameworks such as PyTorch or TensorFlow, which allow for the rapid development of optimization problems in computational settings.  However, these existing tools lack support for sparse linear algebra, severely limiting their use settings such as scientific machine-learning (SciML) and numerical PDEs, where the use of sparsity is often a necessity for achieving efficient algorithms.  In this paper, we develop a set of performant kernels that allow for efficient automatic differentiation of expressions with sparse matrix values; fine-grained parallelism is exploited and performance is highlighted on massively-parallel graphics processing units (GPUs).  We also present several applications that are enhanced by the use of these kernels, such as computing optimal transmission conditions for optimized Schwarz methods or finding sparse preconditioners for conjugate gradients.  The use of backpropagation with these sparse operations results in massive speed gains when compared to their dense counterparts, with speed-ups ranging from $3\times$ to $50\times$ (or more), and enabling computation on larger problems than is possible using dense operations.
\end{abstract}

\ccsdesc[500]{Mathematics of computing~Automatic differentiation}
\ccsdesc[500]{Mathematics of computing~Computations on matrices}
\ccsdesc[500]{Mathematics of computing~Mathematical software performance}

\keywords{sparse matrix operations, backpropagation}

\maketitle

\section{Introduction}
The use of sparse linear algebra is extensive in computational science, arising in the numerical solution of partial differential equations (PDEs)~\cite{mgrid_tut,saad2003iterative}, circuit analysis~\cite{microwave1974}, graph neighborhood analysis~\cite{saad2003iterative,george1993graph}, and more.  The ubiquitous use of sparse matrices underscores the need for efficient execution of sparse operations such as sparse matrix-matrix products (SpSpMM) or sparse matrix-vector products (SpMV).  Many specialized algorithms have been developed to take advantage of the sparsity structure of such matrices and the efficient algorithmic design of such kernels has been well-studied for high performance computing architectures using both CPU and GPU processors~\cite{peng2019slu,chow2015ilu,zhao2021sflu,dalton2015spgemm,SpTMv2014,2012_BeDaOl_amggpu,2015_GuGrOl_gpu,2015_DaBaMeOlGa_merge,2018_BiGrOl_nodeawarespmv}.

The composition of sparse operations forms the backbone of many modern numerical methods. Indeed, to achieve maximal speed and optimal memory usage, the effective use of sparse kernels is often a necessity.  However, optimal performance of the overarching numerical method often requires further parameter tuning and additional setup procedures.  For example, when designing complex multigrid methods, local Fourier analysis (LFA) is often employed to aid in the construction of effective relaxation schemes and the coarse-grid and interpolation operators~\cite{wienandspfa2004,thompsonlfa2021,GeneticMultigrid2003,rodrigolfa2019, PFarrell_etal_2019a}.  In optimized Schwarz methods, heuristics are often employed to obtain optimal transmission values across subdomain boundaries~\cite{taghibakhshi2022mloras,gander2012robin}.  For finite element meshing, small-scale optimizations are often used to generate meshes that give high solution accuracy~\cite{knupp2000meshopt}.  These examples highlight scenarios where a pre-existing setup heuristic is needed. Moreover, there is increased use of robust optimization approaches or machine learning algorithms to improve performance by automating the parameter selection process~\cite{taghibakhshi2022mloras,GreenfeldInterpolation2019,JBrown_etal_2019a,Huang2021LearningRelaxation}.

Modern machine learning frameworks, such as PyTorch~\cite{pytorch}, TensorFlow~\cite{tensorflow}, or Jax~\cite{jax2018github} take advantage of the concept of automatic differentiation~\cite{NolanAutodiff1953,baydinAutodiffsurvey2015,Gebremedhin2020ADSurvey}, allowing rapid development of models and optimization problems without requiring the derivation of analytical gradients.  These libraries allow taking the gradient of complex expressions by decomposing them into small, atomic operations and then linking them together by copious usage of the chain rule.  The work of \citet{taghibakhshi2022mloras}, for example, shows impressive results optimizing neural networks for learning setup parameters for linear solvers, outperforming and generalizing traditional methods~\cite{gander2012robin} that use analytical techniques or heuristics for parameter selection.

However, these frameworks have mostly only implemented automatic differentiation for dense linear algebra.  \citet{taghibakhshi2022mloras}, for example, are limited in the size of the training problems they can employ due to their reliance on dense matrices.  Indeed, dense linear algebra quickly becomes intractable as problem sizes increase.  In this paper, we develop a set of special-purpose sparse kernels to implement automatic differentiation support for compressed sparse row (CSR) matrices. In addition, we demonstrate several novel examples of optimization problems to motivate the new types of computations that can be enabled by our kernels.

The main contributions in this work are:
\begin{enumerate}
\item we develop reverse mode gradients for several sparse operations and detail their implementation on both CPU and GPU (CUDA) processors in \cref{sec:spops};
\item we introduce several optimization problems in \cref{sec:examples}, including sparse linear solver algorithms and graph neural networks~\cite{Wu2021GraphnetSurvey,kipf2017gcn}, to motivate the use of differentiable sparse kernels; and
\item we highlight the performance of the problems and kernels, underscoring the efficiency and speedup gained by using sparse operations.
\end{enumerate}

The full source code is available as an open source implementation of these sparse kernels in PyTorch and can be found at \url{https://github.com/nicknytko/numml}.
\section{Background}
In this section, we introduce notation and motivate later sections of this work
by providing background on the key aspects of algorithmic differentiation, and illustrate
their use through a simple example that includes sparse linear algebra.
\subsection{Chain Rule}\label{subsec:chain}
Given vector $\vec{t} \in \mathbb{R}^{n_t}$ and smooth functions $\vec{x}: \mathbb{R}^{n_t} \to \mathbb{R}^{n_x}$ and $f: \mathbb{R}^{n_x} \to \mathbb{R}$,
we recall the \textit{chain rule} for computing the
partial derivative $\frac{\partial z}{\partial t_i}$ for $z=f(\vec{x}(\vec{t}))$ as
\begin{equation}\label{eqn:chainrule}
    \frac{\partial z}{\partial t_i} = \sum_{j=1}^{n_x} \frac{\partial f}{\partial x_j} \frac{\partial x_j}{\partial t_i}.
\end{equation}
\Cref{eqn:chainrule} extends to arbitrary matrix-valued functions (and, in general, to higher-order tensors as well) by defining the bijective vectorization operator that unwraps matrices to vector form~---~the notation is summarized in~\Cref{def:vec}.
\begin{definition}\label{def:vec}
  Let $\mat{A} \in \mathbb{R}^{m \times n}$.  The \emph{vectorization} operator
  $\vectorize(\cdot): \mathbb{R}^{m \times n} \to \mathbb{R}^{mn}$ is given by
  \begin{equation}
    \vectorize(\mat{A}) = [
    A_{1,1}, A_{2,1}, \ldots, A_{m,1},
    A_{1,2}, A_{2,2}, \ldots, A_{m,2}, \ldots
    A_{1,n}, A_{2,n}, \ldots, A_{m,n}]^T.
  \end{equation}
  That is, $\vectorize(\mat{A})$ is the column-wise form of the matrix as a vector.  The vectorization
  also admits an inverse
 $\vectorize^{-1}(\cdot)$ that unravels the vector form back into the matrix representation.  Moreover, this defines a bijection, $\phi$, that maps indices of elements from the original matrix form to indices on the vectorized form.
\end{definition}
With vectorization, we turn to the chain rule for matrix-valued functions.
First, consider matrices $\mat{T} \in \mathbb{R}^{J_1 \times J_2}$ and $\mat{X}(\mat{T}) \in \mathbb{R}^{I_1 \times I_2}$, and the vector representations
$\vec{t} := \vectorize(\mat{T})$ and $\vec{x} = \vec{x}(\vec{t}) := \vectorize(\mat{X}(\mat{T}))$
with associated index maps $\phi_{\mat{T}}$ and $\phi_{\mat{X}}$.
With indices $j_1 \in \{1,\ldots,J_1\}$ and $j_2 \in \{1,\ldots,J_2\}$, let $j \in \{1,\ldots,J_1 J_2\}$ be the associated
vector index for $\vec{t}$ with $j = \phi_{\mat{T}}(j_1, j_2)$ and let $i$ be the vector index for $\vec{x}$, similarly defined.  With this notation,
$T_{j_1, j_2} = \vectorize({\mat{T}})_j$ and
$X(\mat{T})_{i_1, i_2} = \vectorize({\mat{X}(\mat{T})})_i$.

Given a smooth function $f : \mathbb{R}^{I_1 \times I_2} \to \mathbb{R}$
and the vector forms $\vec{t}$ and $\vec{x}(\vec{t})$,
\Cref{eqn:chainrule} gives
the partial derivative $\frac{\partial z}{\partial T_{j_1,j_2}}$ for $z=f(\mat{X}(\mat{T}))$ as
\begin{equation}
  \frac{\partial z}{\partial t_j} = \sum_{i=1}^{I_1 I_2} \frac{\partial f}{\partial x_i} \frac{\partial x_i}{\partial t_j},
\end{equation}
or equivalently
\begin{equation}
  \frac{\partial z}{\partial (\vectorize{\mat{T}})_j} = \sum_{i=1}^{I_1 I_2} \frac{\partial f}{\partial (\vectorize{\mat{X}})_i} \frac{\partial (\vectorize{\mat{X}})_i}{\partial (\vectorize{\mat{T}})_j}.
\end{equation}
Then, because $\phi_{\mat{X}}$ and $\phi_{\mat{T}}$ are bijections,
\begin{align}
  \frac{\partial z}{\partial T_{j_1,j_2}}
&= \sum_{i=1}^{I_1 I_2} \frac{\partial f}{\partial x_{\phi_{\mat{X}}^{-1}(i)}} \frac{\partial x_{\phi_{\mat{X}}^{-1}(i)}}{T_{j_1,j_2}},\\
\intertext{and the following becomes the \textit{matrix}-valued form of~\Cref{eqn:chainrule}:}
  \frac{\partial z}{\partial T_{j_1,j_2}}
&= \sum_{i_1=1}^{I_1} \sum_{i_2=1}^{I_2} \frac{\partial z}{\partial X_{i_1, i_2}} \frac{\partial X_{i_1, i_2}}{\partial T_{j_1, j_2}}.\label{eqn:tensorchainrule}
\end{align}
It is important to note that this form generalizes to tensor-valued functions as well.

From \Cref{eqn:tensorchainrule}, the first term of the
summation is called the \textit{generalized gradient}:
\begin{equation}\label{eqn:gradfn}
  [\gradfn{\mat{X}}{z}]_{i_1,i_2} = \frac{\partial z}{\partial X_{i_1,i_2}}.
\end{equation}
This represents \textit{the gradient of $z$ with respect to $\mat{X}$}, and
attains the same shape (dimensionality and size in each
dimension) as $\mat{X}$ itself.

Likewise, from \Cref{eqn:tensorchainrule}, the second term in the summation is referred to as the \textit{generalized Jacobian}:
\begin{equation}\label{eqn:jacfn}
  [\jac{\mat{T}}{\mat{X}}]_{(i_1, i_2), (j_1, j_2)} = \frac{\partial X_{i_1, i_2}}{\partial T_{j_1, j_2}},
\end{equation}
where the parentheses in the indexing are used only to emphasize that the first two indices are used for the input, while the last two indices are used for the output; we can view this as either a 4-tensor or as a flattened matrix. From this, we have that \Cref{eqn:tensorchainrule} can be alternatively denoted as the tensor contraction
\begin{equation}
  \gradfn{\mat{X}}{z}^T J_{\mat{T}}(\mat{X}),
\end{equation}
over the indices of the input, $i_1$ and $i_2$.

\subsection{Reverse Mode Automatic Differentiation}\label{subsec:autograd}

Contemporary machine learning frameworks use \textit{reverse-mode automatic differentiation} to compute gradient information, which is an efficient means of computing gradients of scalar-valued functions with respect to tensor-valued inputs~\cite{baydinAutodiffsurvey2015}.  The computation, in essence, is normally executed first as a \textit{forward pass}, with compositions of elementary functions being recorded into a \textit{computation graph}.  A second \textit{backward pass} is then executed, tracing the computation graph backwards from the scalar output back to each input node and computing intermediate gradients at each step.

\begin{figure}
  \centering
  \tikzset{every picture/.style={>=latex,line width=0.5pt}}
\begin{tikzpicture}[x=120pt, y=120pt]
\node [circle, draw] at (0,  0.4) (x) {$\vec{x}$};
\node [circle, draw] at (0,  0.0) (A) {$\mat{A}$};
\node [circle, draw] at (0.6, -0.4) (y) {$\vec{y}$};

\node [circle, draw, label={\small $(\vec{x}^T\mat{A})$}] at (0.5, 0.0) (mv) {MV};
\node [circle, draw, label={\small $(\vec{x}^T\mat{A})\vec{y}$}] at (0.9, 0.0) (dot) {$\langle\cdot,\cdot\rangle$};
\node [circle, draw, label={\small $\sin(\vec{x}^T\mat{A}\vec{y})$}] at (1.4, 0.0) (sin) {$\sin(\cdot)$};
\node [circle, draw, label={\small $2\sin(\vec{x}^T\mat{A}\vec{y})$}] at (1.9, 0.0) (mult) {$\times$};
\node [circle, draw] at (1.6, -0.4) (2) {$2$};
\node [circle, draw] at (2.3, 0.0) (z) {$z$};

\draw[->] (x) to (mv);
\draw[->] (A) to (mv);
\draw[->] (mv) to (dot);
\draw[->] (y) to (dot);
\draw[->] (dot) to (sin);
\draw[->] (sin) to (mult);
\draw[->] (2) to (mult);
\draw[->] (mult) to (z);
\end{tikzpicture}

  \caption{Example computation graph corresponding to evaluating \cref{eqn:ex_fwd}.
    MV denotes matrix-vector multiplication, $\langle\cdot,\cdot\rangle$ an inner product,
    and $\times$ scalar multiplication.  The nodes $\vec{x}$ and $\vec{y}$ are the inputs to
    the expression, while $z$ is the output.
  }\label{fig:comp_graph}
  \Description{A computation graph of \Cref{eqn:ex_fwd}.  The full computation is pulled apart into separate sub-computations.  Arrows show the direction of data flow and order between computations.}
\end{figure}
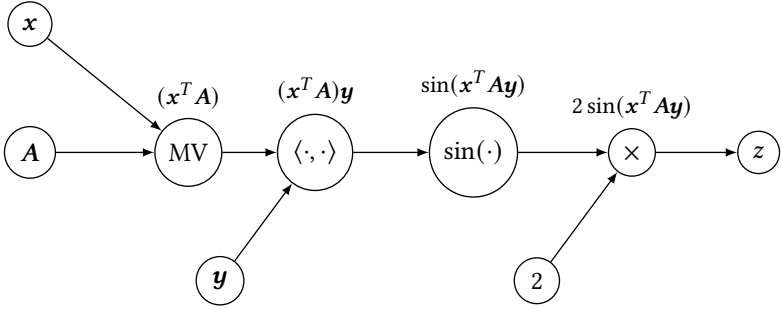

To illustrate, consider the function
\begin{equation}\label{eqn:ex_fwd}
  f(\vec{x}, \vec{y}) = 2 \sin(\vec{x}^T\mat{A}\vec{y}),
\end{equation}
for vectors $\vec{x}\in\mathbb{R}^n$, $\vec{y} \in \mathbb{R}^m$ and matrix $\mat{A} \in \mathbb{R}^{n \times m}$.
We seek the derivative with respect to vectors $\vec{x}$ and $\vec{y}$.
At each node of the computation graph (see \cref{fig:comp_graph}), the gradient of the output is calculated with respect to the input of that particular node; this is then passed further back in the graph.  Formally, letting $z$ be the scalar output of the computation such that $z=f(\vec{x}, \vec{y})$ and $\mat{y} = f_i(\mat{X})$ be the intermediate computation done at node $i$, we find the intermediate gradient with the contraction
\begin{equation}
  \gradfn{\mat{x}}{z} = \gradfn{\mat{y}}{z}^T \jac{f_i(\mat{x})}{\mat{x}}. \label{eqn:contraction}
\end{equation}
This operation is referred to as the \textit{vector-Jacobian product} (VJP) in automatic differentiation.

\section{Kernel Implementations}\label{sec:spops}

In this section, we detail the
forward and backward passes (VJP implementations) for various sparse operations (see \cref{tab:vjp}) with an underlying CSR data structure.
We elect to use the CSR format because it is widely used in general computational settings and has good performance for matrix products, although there may be operations for which other formats are more performant.
\begin{table}
  \centering
  \begin{small}
    \begin{tabular}{llrr}
      \toprule
      \textsc{Method} & \textsc{Operation} & \textsc{WRT} & \textsc{VJP} \\
      \midrule
      \multirow{2}{*}{\textsc{SpMV}}
                      & \multirow{2}{*}{$\mat{A}\vec{x}$}
                                       & $\mat{A}$ & $\vec{v} \vec{x}^T \odot \mask(\mat{A})$ \\
                    &                  & $\mat{x}$ & $\mat{A}^T \vec{v}$ \\
      \midrule
      \multirow{2}{*}{\textsc{SpSpMM}}
                    & \multirow{2}{*}{$\mat{A}\mat{B}$}
                                         & $\mat{A}$ & $(\mat{V}\mat{B}^T) \odot \mask(\mat{A})$ \\
                      &                  & $\mat{B}$ & $(\mat{A}^T \mat{V}) \odot \mask(\mat{B})$ \\
      \midrule
      \multirow{2}{*}{\textsc{SpDMM}}
                      & \multirow{2}{*}{$\mat{A}\mat{B}$}
                                        & $\mat{A}$ & $(\mat{V}\mat{B}^T) \odot \mask(\mat{A})$ \\
                     &                  & $\mat{B}$ & $\mat{A}^T \mat{V}$ \\
      \midrule
      \multirow{2}{*}{\textsc{Sp + Sp}}
                     & \multirow{2}{*}{$\alpha\mat{A} + \beta\mat{B}$}
                                                        & $\mat{A}$ & $\alpha\mat{V} \odot \mask(\mat{A})$ \\
                       &                                & $\mat{B}$ & $\beta\mat{V} \odot \mask(\mat{B})$ \\
      \midrule
      \multirow{2}{*}{\textsc{SpSolve}}
                       & \multirow{2}{*}{$\vec{x}=\mat{A}^{-1}\vec{b}$}
                                                       & $\mat{A}$ & $-\mat{A}^{-T}\vec{v}\mat{x}^T \odot \mask(\mat{A})$ \\
                       &                               & $\vec{b}$ & $\mat{A}^{-T}\vec{v}$ \\
      \bottomrule
    \end{tabular}
  \end{small}
  \caption{Definitions of vector-Jacobian products for different sparse operations.  The vector in the VJP (denoted by $\vec{v}$ or $\mat{V}$) is the intermediate gradient with respect to the output of the operation when running backpropagation.}\label{tab:vjp}
\end{table}

In this section, we will mainly focus on sparse matrices and dense vectors (where applicable) and pay special attention to their sparsity patterns;
notationally, we  denote the \textit{sparsity mask} of some sparse $\mat{M}$ as a boolean-valued matrix encoding the
position of nonzero entries, where
\begin{equation}
  \big[\mask(\mat{M})\big]_{ij} =
  \begin{cases}
    1 & m_{ij} \neq 0 \\
    0 & \text{otherwise.}
  \end{cases}
\end{equation}
With $\odot$ as the standard Hadamard (componentwise) product, $\mat{A}\odot\mask(\mat{M})$ then represents
(sparse) matrix $\mat{A}$ masked to the sparsity of $\mat{M}$.
The sparsity mask also leads to the identity
$\mat{M} \odot \mask(\mat{M}) = \mat{M}$, and, by convention, if $\mat{M}$ is dense then $\mask(\mat{M})$ is also dense.

\begin{figure}
  \centering
  \definecolor{tab-blue}{HTML}{1f77b4}
\definecolor{tab-orange}{HTML}{ff7f0e}
\definecolor{tab-green}{HTML}{2ca02c}
\definecolor{tab-red}{HTML}{d62728}
\definecolor{tab-purple}{HTML}{9467bd}
\definecolor{tab-brown}{HTML}{8c564b}
\definecolor{tab-pink}{HTML}{e377c2}
\definecolor{tab-gray}{HTML}{7f7f7f}
\definecolor{tab-olive}{HTML}{bcbd22}
\definecolor{tab-cyan}{HTML}{17becf}
\begin{tikzpicture}[x=12pt,y=12pt]   %
  \begin{scope}[shift={(20pt, 105pt)}] %

  \def\h{0.25}
  \def\s{0.5}
  \draw[line width=1.0pt] (\s+\h, 0.05) -- (\s, 0.05) -- (\s, -9.25) -- (\s+\h, -9.25); %

  \def\h{-0.25}
  \def\s{10.3}
  \draw[line width=1.0pt] (\s+\h, 0.05) -- (\s, 0.05) -- (\s, -9.25) -- (\s+\h, -9.25); %

  \def\h{0.8}
  \foreach \row/\col in {1/1, 1/2,      1/4,
                         2/1, 2/2, 2/3,      2/5,
                              3/2, 3/3,           3/6,
                         4/1,           4/4, 4/5,      4/7,
                              5/2,      5/4, 5/5, 5/6,      5/8,
                                   6/3,      6/5, 6/6,           6/9,
                                        7/4,           7/7, 7/8,
                                             8/5,      8/7, 8/8, 8/9,
                                                  9/6,      9/8, 9/9
                   }
     {\fill[fill=tab-blue] (\col,-\row) rectangle ++(\h,\h);} %

  \def\s{11.0}
  \def\h{0.8}
  \foreach \row/\col in {1/1,
                         2/1,
                              3/2,
                         4/1,
                              5/2,
                              6/2,
                                   7/3,
                                   8/3,
                                   9/3
                   }
     {\fill[fill=tab-red] (\s+\col,-\row) rectangle ++(\h,\h);}

  \def\h{0.25}
  \def\s{11.5}
  \draw[line width=1.0pt] (\s+\h, 0.05) -- (\s, 0.05) -- (\s, -9.25) -- (\s+\h, -9.25); %

  \def\h{-0.25}
  \def\s{15.3}
  \draw[line width=1.0pt] (\s+\h, 0.05) -- (\s, 0.05) -- (\s, -9.25) -- (\s+\h, -9.25); %

  \def\h{0.25}
  \def\s{16.5}
  \draw[line width=1.0pt] (\s+\h, 0.05) -- (\s, 0.05) -- (\s, -9.25) -- (\s+\h, -9.25);

  \def\h{-0.25}
  \def\s{20.3}
  \draw[line width=1.0pt] (\s+\h, 0.05) -- (\s, 0.05) -- (\s, -9.25) -- (\s+\h, -9.25);

  \def\s{16.0}
  \def\h{0.8}
  \foreach \row/\col in {1/1,
                         2/1, 2/2,
                         3/1, 3/2,
                         4/1, 4/2, 4/3,
                         5/1, 5/2, 5/3,
                              6/2, 6/3,
                         7/1,      7/3,
                              8/2, 8/3,
                              9/2, 9/3
                   }
    {\fill[fill=tab-green] (\s+\col,-\row) rectangle ++(\h,\h);} %

  \node at (10.9, -4.5) {$\times$};
  \node at (15.9, -4.5) {$=$};

  \draw[-{Latex[]}, line width=1.2pt] (17.4, -0.6) parabola bend(9.4, 0.5) (1.4, -0.6);
  \draw[-{Latex[]}, line width=1.2pt] (17.4, -0.6) parabola bend(9.4, 0.5) (2.4, -0.6);
  \draw[-{Latex[]}, line width=1.2pt] (17.4, -0.6) parabola bend(9.4, 0.5) (4.4, -0.6);
  \draw[-{Latex[]}, line width=1.2pt] (17.4, -0.6) parabola bend(14.9, -0.2) (12.4, -0.6);
  \draw[-{Latex[]}, line width=1.2pt] (17.4, -0.6) parabola (12.4, -1.6);
  \draw[-{Latex[]}, line width=1.2pt] (17.4, -0.6) parabola (12.4, -3.6);

  \draw[-{Latex[]}, line width=1.2pt] (19.4, -7.6) parabola bend(13.4, -6.0) (7.4, -7.6);
  \draw[-{Latex[]}, line width=1.2pt] (19.4, -7.6) parabola bend(13.4, -6.0) (8.4, -7.6);
  \draw[-{Latex[]}, line width=1.2pt] (19.4, -7.6) parabola bend(13.4, -6.0) (9.4, -7.6);
  \draw[-{Latex[]}, line width=1.2pt] (19.4, -7.6) parabola[bend at end] (14.4, -6.6);
  \draw[-{Latex[]}, line width=1.2pt] (19.4, -7.6) parabola bend(16.9, -7.4) (14.4, -7.6);
  \draw[-{Latex[]}, line width=1.2pt] (19.4, -7.6) parabola (14.4, -8.6);

  \end{scope}
\end{tikzpicture}
  \caption{Back-propagation through a sparse matrix-matrix multiplication.  The arrows demonstrate how gradient information flows from nonzero entries in the output to nonzero entries in the input.}\label{fig:sparse_back_prop}
  \Description{A visual depiction of a sparse, square matrix multiplying a sparse, rectangular matrix.  On the right is the output of the matrix-matrix multiply.  Arrows point from entries in the output to entries in the two input matrices, denoting the flow of gradient information to nonzeros.}
\end{figure}
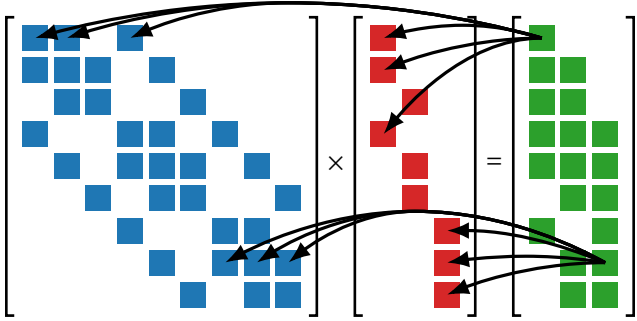

We propagate \textit{sparse} gradients (\cref{fig:sparse_back_prop}) whereby functions that take sparse inputs will have gradients whose sparsity mask will match those of the inputs.  This trade-off leads to much smaller memory and computational costs (as observed indirectly in \cref{tab:timing_results_blascupy,tab:timing_results_sp_vs_dense}) as, during optimization, only nonzero entries in the inputs will receive gradients.  For example, optimizing a problem with a sparse-matrix input and scalar output will only lead to optimization of the nonzero entries of the input.  In contrast, if the dense gradients (where zero entries of the input matrices contribute to the gradient) are propagated, then any benefits of the original sparse data structure would be lost.

We next detail the forward and backward implementation (also known as VJP) for several sparse operations.  For derivations of the matrix reverse-mode updates, we point the reader to~\cite{MatAutodiff2008} and remark that with  minor consideration of the sparsity patterns, the results extend to sparse matrices.  See \cref{sec:spop_timing} for timing results that show the speedup over the dense representation of these operations.
\subsection{Sparse Methods}
Ordinarily, computing the vector-Jacobian products for dense matrix operations will give dense intermediate gradients.  However, computing and storing dense gradients is not scalable for large matrix sizes, both in terms of the necessary memory and floating point operations required.  Therefore, we elect to keep intermediate gradients with the same sparsity mask as their respective element.  For example,
\begin{equation}
  \mask(\gradfn{\mat{X}}{z}) = \mask(\mat{X}),
\end{equation}
for scalar $z$ that is a function of matrix $\mat{X}$.  In the derivation of the VJP, we do this by masking the output (removing entries that do not agree) with the sparsity mask of the respective variable.  In practice, knowing the output sparsity mask allows us to skip unnecessary computations by only focusing on matrix entries that we know a priori to be nonzero; the nonzero pattern of the gradient is effectively constrained to match that of its variable.
\subsubsection{Sparse Matrix-Vector (SpMV)}\label{subsec:spmv}
The sparse matrix-vector product computes
\begin{equation}
\vec{y} \gets \mat{A} \vec{x}
\end{equation}
for sparse $\mat{A} \in \mathbb{R}^{m \times n}$ and dense $\vec{x} \in \mathbb{R}^{n}$, and outputs dense $\vec{y} \in \mathbb{R}^{m}$.  The CSR representation allows for direct evaluation of inner products of each row of $\mat{A}$ and $\vec{x}$, and the inner products can be executed in parallel on a GPU\@.

For the backward pass, we compute $\mat{A}^T\vec{v}$ (resp.\ $\vec{v}\vec{x}^T$) for the gradient with respect to $\vec{x}$ (resp.\ $\mat{A}$), for intermediate gradient value $\vec{v}$.  This does not directly align with the sparse structure. Hence, we use the reduction-based algorithm as outlined in~\citet{SpTMv2014}: inner products of the rows of $\mat{A}$ and $\vec{v}$ are computed and atomically reduced into correct entries in the output vector.  Computing $\vec{v}\vec{x}^T$ is easily parallelized, since it is the outer product between two dense vectors that is masked to a sparse matrix, only requiring computation of nonzero entries of $\mat{A}$.
\subsubsection{Sparse-Sparse Matrix Multiply (SpSpMM)}\label{subsec:spspmm}
The sparse-sparse matrix multiplication primitive is given as
\begin{equation}
\mat{C} \gets \mat{A}\mat{B},
\end{equation}
where $\mat{A} \in \mathbb{R}^{m \times n}$, $\mat{B} \in \mathbb{R}^{n \times p}$, and $\mat{C} \in \mathbb{R}^{m \times p}$, with all three matrices stored in CSR format.  In the forward pass, we follow the parallel SpGEMM algorithm from~\citet{dalton2015spgemm}, where we compute intermediate products of each row of $\mat{A}$ and the entirety of $\mat{B}$, then reduce and collect redundant entries to form $\mat{C}$.

For the backward pass, we compute $(\mat{V}\mat{B}^T) \odot \mask(\mat{A})$ and $(\mat{A}^T\mat{V}) \odot \mask(\mat{B})$.  Noting that the product $\mat{V}\mat{B}^T$ is the inner product between rows of both $\mat{V}$ and $\mat{B}$, our data accesses aligns directly with the CSR structure.  Computing $\mat{A}^T\mat{V}$, however, accesses columns of both $\mat{A}$ and $\mat{V}$; we thus follow a modified version of the SpGEMM algorithm from above that operates on columns of $\mat{A}$ instead.  Gradients propagate only for nonzero entries in common in the rows of $\mat{A}$ and columns of $\mat{B}$ (see \cref{fig:sparse_back_prop}).
\subsubsection{Sparse-Dense Matrix Multiply (SpDMM)}\label{subsec:spdmm}
For the SpDMM routine, we compute
\begin{equation}
\mat{C} \gets \mat{A}\mat{B}
\end{equation}
as in the SpSpMM case except for $\mat{B}$ and $\mat{C}$ now being dense.  In the forward pass, we parallelize with each fine-grained operation focusing on an entry of $\mat{C}$.  That is, we compute the respective inner product between a row of $\mat{A}$ and a column of $\mat{B}$ which, because of its dense structure, does not incur any significant penalties for column accesses.

On the backward pass, we compute $(\mat{VB}^T) \odot \mask(\mat{A})$ (as before) and $\mat{A}^T\mat{V}$.  For the latter, we take the sparse transpose of $\mat{A}$ and re-execute the forward routine to compute the gradient.
\subsubsection{Sparse + Sparse}\label{subsec:spadd}
In a sparse add, we compute the linear combination of two sparse matrices as in
\begin{equation}
  \mat{C} \gets \alpha \mat{A} + \beta \mat{B},
\end{equation}
which we write in a general form so that both $\mat{A} + \mat{B}$ and $\mat{A} - \mat{B}$ are computed with the same method.  A key observation is that (with slight abuse of notation),
\begin{equation}
  \mask(\mat{C}) = \mask(\mat{A}) \cup \mask(\mat{B}),
\end{equation}
meaning the computation of $\mat{C}$ is viewed as a union over the rows of $\mat{A}$ and $\mat{B}$.  Moreover, this form is implemented in parallel over each row.

To compute the backward pass, we consider the gradient with respect to $\mat{A}$, $\mat{B}$ as $\alpha \mat{V} \odot \mask(\mat{A})$, $\beta \mat{V} \odot \mask(\mat{B})$, respectively.  Each are found as the row-wise reduction from $\mat{V}$ to the sparsity mask of $\mat{A}$ or $\mat{B}$. Because both $\mask(\mat{A})$, $\mask(\mat{B}) \subseteq \mask(\mat{V})$, we need only to compute matching nonzero entries in both matrices, which can again be implemented in parallel over the rows.
\subsubsection{Sparse Triangular Solve}\label{subsec:sptrsv}
A sparse triangular solve is an operation to compute the value of $\vec{x}$ in the matrix equation
\begin{equation}
  \mat{L}\vec{x} = \vec{b},
\end{equation}
where $\mat{L}$ has a lower triangular form, i.e. $\mat{L}_{ij} \neq 0$ if $i \geq j$.  Without loss of generality, we  consider upper-triangular systems $\mat{U}$ using matrix flip operations to convert $\mat{U}\vec{x} = \vec{b}$ into an equivalent lower-triangular system.
Such a system has a (relatively) simple routine for computing the linear solve:
each row depends on the intermediate values of previous rows only and no
intermediate preprocessing is needed.

For the forward pass, we use the synchronization-free GPU triangular solve detailed in~\citet{Capellini2020} to exploit the limited parallelism that may exist in computing $\vec{x}$.
On the backward pass, we can refer to the general VJP rule for a sparse linear solve; we seek $\mat{L}^{-T}\vec{v}$ and $-\mat{L}^{-T}\vec{vx}^T \odot \mask(\mat{L})$ for gradients with respect to $\vec{b}$ and $\mat{L}$, respectively.  We first find $\mat{L}^{-T}\vec{v}$ with our existing forward triangular solve routine. We then observe that the gradient with respect to $\mat{L}$ contains the $\vec{b}$ gradient term as a masked outer product, so we can re-use the result.  The masked outer-product can be executed in parallel over the nonzero entries of $\mat{L}$.
\subsubsection{Sparse Direct Solve}\label{subsec:spsolve}
For a sparse direct solve, we seek the solution to
\begin{equation}
  \mat{A}\vec{x} = \vec{b},
\end{equation}
for $\vec{x}$, for any square, nonsingular $\mat{A}$.  A standard way to compute this linear solve is to first decompose $\mat{A}$ as the product
\begin{equation}
  \mat{P}\mat{A}\mat{Q} = \mat{L}\mat{U},
\end{equation}
where $\mat{L}$ and $\mat{U}$ are sparse lower- and upper-triangular systems, and $\mat{P}$, $\mat{Q}$ are row and column permutation matrices, respectively, that aim to reduce the amount of extra nonzero entries generated (fill) in the $\mat{L}$ and $\mat{U}$ factors.  We then find $\vec{x}$ by solving the triangular systems
\begin{align}
  \mat{L}\vec{y} &= \mat{P} \vec{b}, \label{eqn:spsolve_L_solve} \\
  \mat{U}\bar{\vec{x}} &= \vec{y}, \label{eqn:spsolve_U_solve} \\
  \vec{x} &= \mat{Q}\bar{\vec{x}},
\end{align}
where $\vec{y}, \bar{\vec{x}}$ are intermediate vectors used in the computation.

To find the factorization, we use the existing SuperLU package~\cite{Demmel1999} that gives the $\mat{L}$ and $\mat{U}$ factors, as well as the row and column permutations using an approximate minimum degree algorithm to reduce fill.

For the forward pass, the intermediate factors $\mat{L}$ and $\mat{U}$ are
constructed, followed by the respective triangular solves in
\Cref{eqn:spsolve_L_solve,eqn:spsolve_U_solve}.  Since we only expose the
entire action of computing the linear solve, we do not return the $\mat{L}$ and
$\mat{U}$ factors; this avoids taking the intermediate gradient with respect to
each.  We remark that this can be nontrivial to implement while keeping
intermediate computations sparse and, thus, leave this extension to a future study.

For the backward pass, we compute $\mat{A}^{-T}\vec{v}$ and
$-\mat{A}^{-T}\vec{vx}^T \odot \mask(\mat{A})$ for the gradients with respect
to $\vec{b}$ and $\mat{A}$, respectively.  We reuse the existing $\mat{L}$
and $\mat{U}$ that we found in the forward pass, as in
\begin{equation}
  \mat{A}^T = \big(\mat{P}^T\mat{L}\mat{U}\mat{Q}^T\big)^T = \mat{Q}\mat{U}^T\mat{L}^T\mat{P}.
\end{equation}
Then, defining $\bar{\vec{w}}=\mat{P}\vec{w}$, we compute $\vec{w} = \mat{A}^{-T}\vec{v}$ by solving in order
\begin{align}
  \mat{U}^T\vec{y} &= \mat{Q}^T\vec{v}, \\
  \mat{L}^T\bar{\vec{w}} &= \vec{y}, \\
  \vec{w} &= \mat{P}^T\bar{\vec{w}},
\end{align}
which requires two triangular solves and two vector permutations.  We next find the gradient with respect to $\mat{A}$ in a similar fashion to the triangular solve, by computing the gradient as the masked outer-product of $\vec{w}$ and $\vec{v}$.
\subsection{Timings}\label{sec:spop_timing}
We present timing results for our implementation of the sparse kernels outlined above
along with a comparison of CPU and GPU (CUDA) timings in
\cref{tab:timing_results_blascupy,tab:timing_results_sp_vs_dense}.  For comparison, we have also included the respective
\textit{dense} operation in applicable cases, such as the dense
matrix-vector (DMV) or dense-dense matrix-matrix product (DDMM).  These dense
implementations use the PyTorch built-in matrix routines.  We also compare
with NumPy/SciPy CPU and performant CUDA versions (through CuPy) for the forward
passes.  For the CuPy tests, we ran an extra warmup iteration to allow for the
just-in-time kernels to be compiled.

\begin{table*}
  \centering
  \newcommand{\FF}{F$\to$}
\newcommand{\BB}{B$\gets$}
\newcommand{\x}{{$\times$}}

\begin{tabular}{
  @{}
  l %
  l %
  S[table-format=4, table-number-alignment = right, table-text-alignment = right] %
  |
  S[table-auto-round, table-format=4.2] %
  S[table-auto-round, table-format=4.2] %
  |
  S[table-auto-round, table-format=4.2] %
  S[table-auto-round, table-format=4.2] %
  |
  S[table-auto-round, table-format=4.1\x, table-number-alignment = left] %
  @{}}
 \toprule
  \multicolumn{2}{c}{\multirow{2}{*}{Test name}} & {\multirow{2}{*}{Iter.}} & \multicolumn{2}{c|}{CPU} & \multicolumn{2}{c|}{GPU} & {Speedup} \\
                                                 & & & {\textbf{Ours}} & {BLAS} & {\textbf{Ours}} & {CuPy} & {\textbf{Ours}} \\
\midrule
\multirow{2}{*}{\textsc{SpMV}}
& \FF{} & 1000 & 0.141 & 0.092 & 0.045 & 0.049 & 3.150\x \\
& \BB{} & 1000 & 0.595 &       & 0.198 &       & 3.003\x \\
\midrule
\multirow{2}{*}{\textsc{SpSpMM}}
& \FF{} & 100 & 0.368 & 0.117 & 0.070 & 0.041 & 5.289\x  \\
& \BB{} & 100 & 4.890 &       & 0.166 &       & 29.490\x \\
\midrule
\multirow{2}{*}{\textsc{SpDMM}}
& \FF{} & 100 & 567.078  & 415.882 & 4.229  & 2.774 & 134.086\x \\
& \BB{} & 100 & 1203.333 &         & 11.188 &       & 107.551\x \\
\midrule
\multirow{2}{*}{\textsc{Sp + Sp}}
& \FF{} & 100 & 0.073 & 0.034 & 0.011 & 0.011 & 6.813\x \\
& \BB{} & 100 & 0.133 &       & 0.025 &       & 5.227\x \\
\midrule
\multirow{2}{*}{\textsc{SpTRSV}}
& \FF{} & 100 & 0.030  & 32.746 & 4.710  & 0.767 & \multicolumn{1}{r}{--\textsuperscript{**}} \\
& \BB{} & 100 & 13.462 &        & 10.997 &       & 1.224\x \\
\bottomrule
\end{tabular}

  {\scriptsize \textsuperscript{**} No speed-up due to insufficient parallelism.}
  \caption[cap:timing_results_blascupy]{
    Timing results comparing the CPU- and CUDA- based implementations against BLAS and CuPy; all units are in seconds.
    These tests were performed on a static matrix size of \num{32768} $\times$ \num{32768}.  The notation \textbf{\FF} refers to the \textbf{F}orward pass, \textbf{\BB} refers to the \textbf{B}ackward pass, \textbf{Iter.} refers to the number of times the specific test was run (times listed are total wall-clock time elapsed), and \textbf{Speedup} is computed as CPU time $/$ GPU time.}\label{tab:timing_results_blascupy}
\end{table*}

\begin{table*}
  \centering
  \newcommand{\FF}{F$\to$}
\newcommand{\BB}{B$\gets$}
\newcommand{\x}{{$\times$}}

\begin{tabular}{
  @{}
  l %
  l %
  S[table-format=4, table-number-alignment = right, table-text-alignment = right] %
  |
  S[table-auto-round, table-format=4.2] %
  S[table-auto-round, table-format=4.2] %
  S[table-auto-round, table-format=4.1\x, table-number-alignment = left] %
  |
  S[table-auto-round, table-format=4.2] %
  S[table-auto-round, table-format=4.2] %
  S[table-auto-round, table-format=4.1\x, table-number-alignment = left] %
  @{}}
 \toprule
  \multicolumn{2}{c}{\multirow{2}{*}{Test name}} & {\multirow{2}{*}{Iter.}} & \multicolumn{3}{c|}{CPU} & \multicolumn{3}{c}{GPU} \\
  & & & {Sp \textbf{Ours}} & {D Torch} & {Speedup} & {Sp \textbf{Ours}} & {D Torch} & {Speedup} \\
\midrule
\multirow{2}{*}{\textsc{Mat-Vec}}
& \FF{} & 1000 & 0.141 & 143.441  & 1024.5714285714\x & 0.045 &  2.985 & 66.3333333333\x \\
& \BB{} & 1000 & 0.595 & 1509.088 & 2515.15\x         & 0.198 & 21.806 & 110.1313131313\x \\
\midrule
\multirow{2}{*}{\textsc{Mat-Mat}$^\dag$}
& \FF{} & 100 & 0.368 & 1455.164 & 3954.25\x & 0.070 &  7.672 & 109.6\x  \\
& \BB{} & 100 & 4.890 & 4306.136 & 880.600\x & 0.166 & 23.257 & 140.102\x \\
\midrule
\multirow{2}{*}{\textsc{Mat add}}
& \FF{} & 100 & 0.073 &  60.978 & 835.315\x & 0.011 & 0.941 & 85.54\x \\
& \BB{} & 100 & 0.133 & 123.519 & 928.714\x & 0.025 & 2.596 & 103.84\x \\
\midrule
\multirow{2}{*}{\textsc{Tri solve}}
& \FF{} & 100 & 0.030  &  35.092 & 1169.73\x & 4.710  & 0.695 & 0.147\x \\
& \BB{} & 100 & 13.462 & 863.779 & 64.1166\x & 10.997 & 3.499 & 0.318\x \\
\bottomrule
\end{tabular}

  {\scriptsize
    $^\dag$ This refers to a sparse-sparse matrix product and dense-dense matrix product for the sparse and dense cases, respectively.}
  \caption[cap:timing_results_sp_vs_dense]{
    Timing results comparing our sparse implementations versus their respective dense counterparts in PyTorch; all units are in seconds.
    These tests were performed on a static matrix size of \num{32768} $\times$ \num{32768}.  The notation \textbf{Sparse} refers to our sparse implementations, dense refers to PyTorch's dense implementations, \FF refers to the \textbf{F}orward pass, \BB refers to the \textbf{B}ackward pass, \textbf{Iter.} refers to the number of times the specific test was run (times listed are total wall-clock time elapsed), and \textbf{Speedup} is computed as dense running time $/$ sparse running time.}\label{tab:timing_results_sp_vs_dense}
\end{table*}

\begin{figure*}
  \centering
    \includegraphics{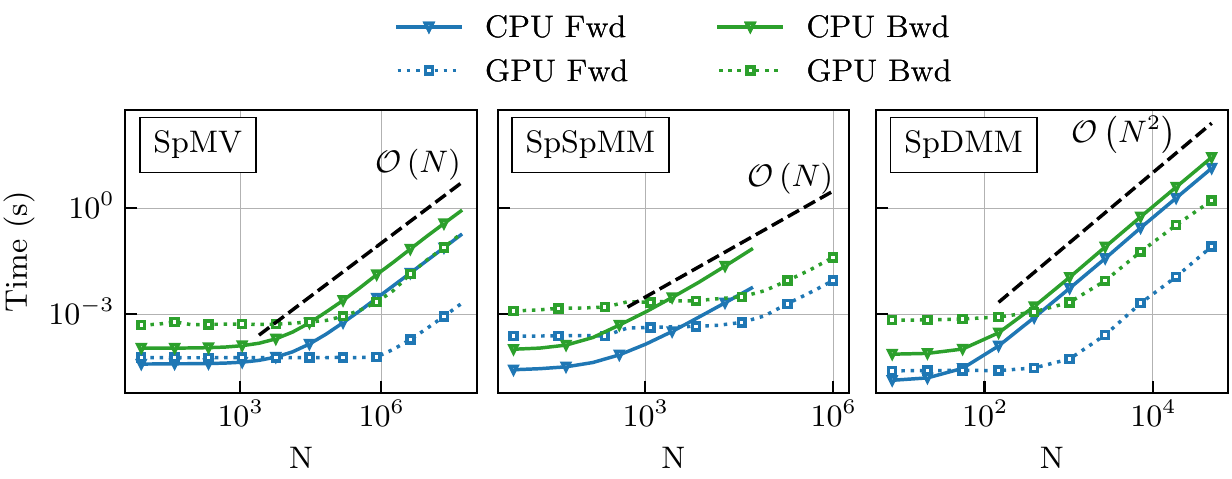}
    \caption{Scaling results for the forward and backward passes on several key sparse operations.  The CPU running times are in blue, while the CUDA runtimes are in green.  Triangle markers denote the backward pass routines.}\label{fig:scalings_all}
    \Description{Scaling plots for SpMV, SpSpMM, and SpDMM\@.  The first two run in linear time in the forward and backward routines, while the latter
      runs in quadratic time for the forward and backward routines.}
\end{figure*}

\Cref{fig:scalings_all} highlights scaling
results for the sparse matrix-vector product, sparse-sparse matrix product, and
sparse-dense matrix product. These show the running time as a factor of the number
of nonzeros in the matrix given in \Cref{eqn:mat_1d_fd}, and indicate that
forward and backward passes for the SpMV and SpSpMM are both linear in the
nonzeros, while the forward and backward passes for the SpDMM are quadratic in
the number of nonzeros.

These timing results are run on a single compute node of the Narval cluster
with two AMD Milan \num{7413} processors and four NVIDIA A100 GPUs, though for our tests
we use only one CPU and one GPU\@. Our CPU-based implementations are single
threaded only and so, for comparison, we limit the number of threads in the
dense operations to one.
\section{Applications}\label{sec:examples}
To highlight the power and ease-of-use (especially on GPU accelerators) of our differentiable sparse kernels, we present several optimization problems based on applications and methods in computational science; these employ heavy use of sparse matrix computations.  These examples could not be easily done before (without careful bespoke sparse matrix routines); with our automatic differentiation framework, however, they can be readily implemented with a small amount of additional effort.

We first define some common test problems.  As an example sparse matrix, we define $\mat{A}_{N} \in \mathbb{R}^{N\times N}$ as
\begin{equation}\label{eqn:mat_1d_fd}
  \mat{A}_{N} = \begin{bmatrix}
              2  &     -1 &        & \\
              -1 & \ddots & \ddots & \\
                 & \ddots & \ddots & -1\\
                 &        &     -1 & 2
            \end{bmatrix},
\end{equation}
which is a standard test problem in computational science and comes from the finite-difference discretization of the 1D Poisson problem~\cite{saad2003iterative,quarteroni2007numerical},
\begin{align}
  -\grad^2 u &= f \quad\text{in $\Omega$}, \\
  u &= 0 \quad\text{on $\partial\Omega$},
\end{align}
for the domain $\Omega=(0, N+1) \subseteq \mathbb{R}$, with $N$ being the number of interior grid points in the discretization~---~the domain is selected to cancel any constant scaling of the matrix.

As a more complex example, we will also use a 2D version of the Poisson problem, defined by tensor products on \Cref{eqn:mat_1d_fd} over the $x$ and $y$ dimensions, as in
\begin{equation}\label{eqn:mat_2d_fd}
  \mat{A}_{N_x \times N_y} = \big(\mat{A}_{N_x} \otimes \mat{I}_{N_y}\big) + \big(\mat{I}_{N_x} \otimes \mat{A}_{N_y}\big),
\end{equation}
where $\otimes$ denotes the standard matrix Kronecker product, and $\mat{A}_{N}$ refers to the one-dimensional discretization with $N$ interior grid points in~\cref{eqn:mat_1d_fd}.  This results in a less trivial sparsity pattern: the matrix is no longer tri-diagonal and is, instead, pentadiagonal.

The remainder of this section describes several problems that involve heavy use of sparse operations.  Corresponding timings for each of these examples are included in each subsection, along with comparisons to timings using only dense operations.
\subsection{Entry-wise Jacobi Relaxation}\label{subsec:jacobi}
The weighted Jacobi method~\cite{saad2003iterative} is an iterative solver for linear systems whereby the inverse matrix is approximated by inverting only the diagonal entries of $\mat{A}$ according to some parameter (weight) $\omega$.  In practice, the best choice of $\omega$ is problem specific and choosing a suboptimal value can lead to a method that converges slowly or even diverges.  Here, we explore how to find $\omega$ automatically by solving an optimization problem.

Consider solving
\begin{equation}\label{eqn:lin_solve_Axb}
\mat{A}\mat{x} = \vec{b},
\end{equation}
where $\mat{A} \in \mathbb{R}^{N \times N}$ is a sparse matrix with diagonal $\mat{D}$.
The weighted Jacobi method is
\begin{equation}
  \vec{x}^{(k+1)} = \omega\mat{D}^{-1}\vec{b} + \big(\mat{I} - \omega\mat{D}^{-1}\mat{A}\big)\vec{x}^{(k)} = \vec{x}^{(k)} + \omega\mat{D}^{-1}\big(\vec{b}-\mat{A}\vec{x}^{(k)}\big).
\end{equation}
The weight value, $\omega$, specifies how much of the correction, $\mat{D}^{-1}\big(\vec{b}-\mat{A}\vec{x}^{(k)}\big)$, is added to the current approximation at each iteration.  For trivial problems, there are known values of $\omega$ that lead to convergent solvers; however, this is not always the case for more difficult problems.

To make the problem more interesting, we will consider an \textit{entrywise} weighting scheme: denoting $\mat{\Omega}$ as the diagonal scaling matrix of the entries of $\vec{x}^{(k)}$, we rewrite the Jacobi iteration as
\begin{equation}
  \vec{x}^{(k+1)} = \mat{\Omega} \mat{D}^{-1} \vec{b} + \big(\mat{I} - \mat{\Omega}\mat{D}^{-1}\mat{A}\big)\vec{x}^{(k)} = \vec{x}^{(k)} + \mat{\Omega}\mat{D}^{-1}\big(\vec{b}-\mat{A}\vec{x}^{(k)}\big).
\end{equation}
\newcommand{\jacobiletter}{g}
To simplify the notation, we write $\vec{\jacobiletter}\big(\vec{x}^{(k)}, \vec{b};\:\mat{\Omega}\big)$ as the function that applies one iteration of the weighted Jacobi scheme to $\vec{x}^{(k)}$ in order to approximate the solution of $\mat{A}\vec{x}=\vec{b}$.

\begin{figure}
  \centering
  \begin{tikzpicture}[remember picture, overlay]
\coordinate (x_shift) at ([xshift=\linewidth]pic cs:jacobi_setup_top);
\coordinate (jacobi_setup_top_rect) at ({pic cs:jacobi_setup_mid} -| x_shift);
\coordinate (jacobi_setup_mid_rect) at ({pic cs:jacobi_setup_bot} -| x_shift);
\path[overlay, fill=red!10] ($(pic cs:jacobi_setup_top) + (-0.1,0.24)$) rectangle ($(jacobi_setup_top_rect) + (-2.5,-0.1)$);
\path[overlay, fill=green!10] ($(pic cs:jacobi_setup_mid) + (-0.1,0.22)$) rectangle ($(jacobi_setup_mid_rect) + (-2.5,0.20)$);
\coordinate (jacobi_spdiag_top_rect) at ({pic cs:jacobi_spdiag_mid} -| x_shift);
\coordinate (jacobi_spdiag_mid_rect) at ({pic cs:jacobi_spdiag_bot} -| x_shift);
\path[overlay, fill=red!10] ($(pic cs:jacobi_spdiag_top) + (-0.1,0.22)$) rectangle ($(jacobi_spdiag_top_rect) + (-2.5,-0.05)$);
\path[overlay, fill=green!10] ($(pic cs:jacobi_spdiag_mid) + (-0.1,0.20)$) rectangle ($(jacobi_spdiag_mid_rect) + (-2.5,0.20)$);
\end{tikzpicture}
\begin{minted}[xleftmargin=4em,escapeinside=||]{Python}
  import torch
  import numml.sparse as sp

  # Problem setup
  N = 16
|\tikzmark{jacobi_setup_top}|- A = (torch.diag(torch.ones(N) * 2) +
-      torch.diag(torch.ones(N - 1)*-1, diagonal=1) +
-      torch.diag(torch.ones(N - 1)*-1, diagonal=-1))
- I = torch.eye(N)
- Dinv = torch.diag(1. / torch.diag(A))
|\tikzmark{jacobi_setup_mid}|+ A = sp.eye(N)*2. - sp.eye(N, k=1) - sp.eye(N, k=-1)
+ I = sp.eye(N)
+ Dinv = sp.diag(1. / A.diagonal())
|\tikzmark{jacobi_setup_bot}|
  omega = torch.ones(N, requires_grad=True)
  optimizer = torch.optim.Adam([omega], lr=1e-2)
  for i in range(100):
      optimizer.zero_grad()
      x = torch.randn(N)
      x /= torch.linalg.norm(x)

      # Convert to a diagonal matrix
|\tikzmark{jacobi_spdiag_top}|-     Omega = torch.diag(omega)
|\tikzmark{jacobi_spdiag_mid}|+     Omega = sp.diag(omega)
|\tikzmark{jacobi_spdiag_bot}|
      x = (I - Omega) @ (Dinv @ A @ x)
      loss = (x@(A@x))
      loss.backward()
      optimizer.step()
\end{minted}
  \caption{Implementation code for the entry-wise weighted Jacobi optimization problem.  Differences between the sparse (green, +) and dense (red, -) implementations are shown with highlighted lines.}\label{fig:snippet_jacobi}
\end{figure}

\begin{figure}
  \centering
  \includegraphics{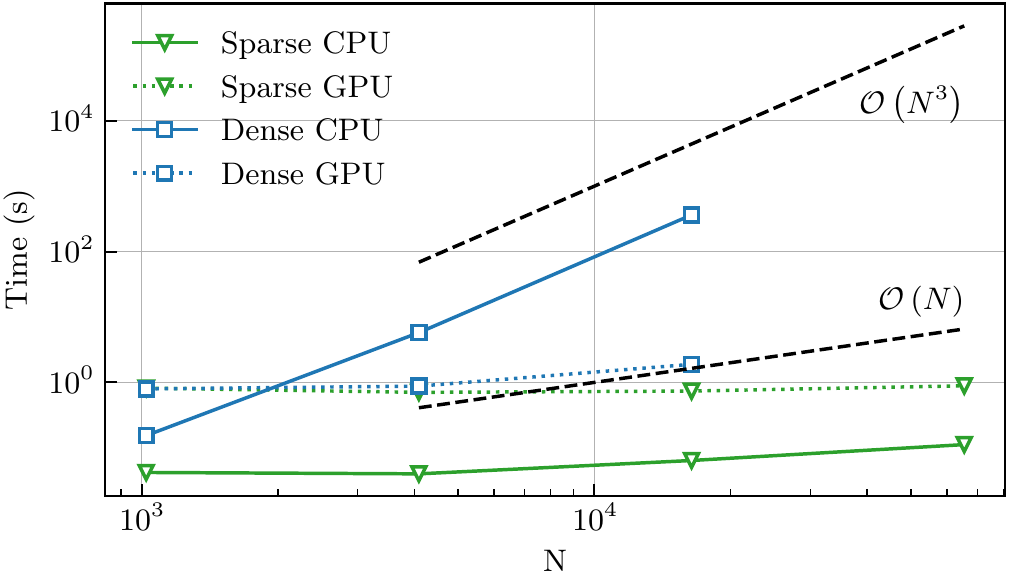}
  \caption{Timing result for optimizing over the entry-wise Jacobi method.  The dense implementation shows obvious super-linear scaling, while the sparse implementation does not yet show asymptotic scaling behavior.}\label{fig:timing_results_jacobi}
\end{figure}

We generate a finite-difference matrix $\mat{A}_N$ from~\Cref{eqn:mat_1d_fd} with $N=16$, then optimize the entries of $\mat{\Omega}$ by minimizing
\begin{equation}
  \ell = \sum_{j=1}^{J} (\vec{\jacobiletter}(\vec{x}_j, \vec{0};\:\mat{\Omega}))^T\mat{A}(\vec{\jacobiletter}(\vec{x}_j, \vec{0};\:\mat{\Omega})),
\end{equation}
where $\{\vec{x}_1, \vec{x}_2, \ldots, \vec{x}_J\}$ is a set of test vectors in $\mathbb{R}^N$ with random unit-normally distributed entries.  This optimizes the one-step Jacobi error reduction when solving with a zero right-hand side.  An annotated code listing that shows the implementation of this problem can be seen in \cref{fig:snippet_jacobi}.  In that figure, we highlight the differences between using our sparse routines and PyTorch's built-in dense operations; changes are only required in defining data types.  We present timing results of one training epoch in \cref{fig:timing_results_jacobi}, for both sparse and dense implementations on CPU and GPU\@.  With our sparse implementation, we have much better scaling as a function of the problem size.  The dense implementation, on the other hand, has roughly cubic scaling.  The dense case is also truncated early (in terms of $N$), as the test application had run out of available memory at that point.

\begin{figure}
  \centering
  \includegraphics{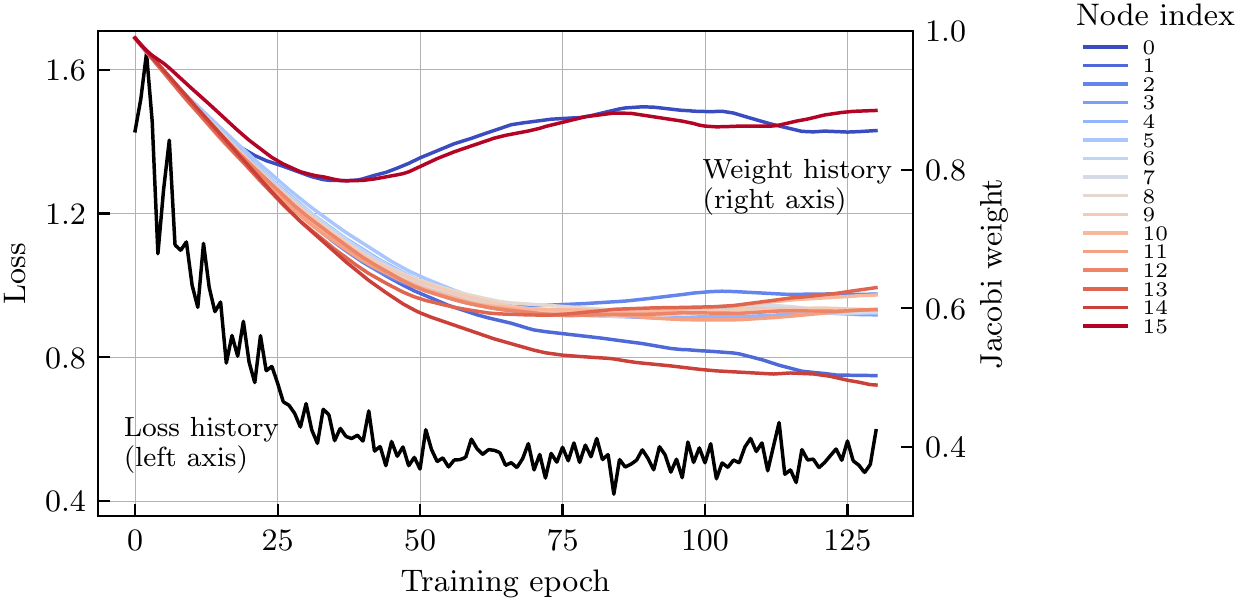}
  \caption{Loss history (black) obtained from optimizing over the Jacobi relaxation weights and
    individual node weights (colored, for each node $0,\ldots,15$).  Node indices are ordered by their $x$ position in the domain, such that node $0$ is the left boundary and $15$ is the right boundary.}\label{fig:jacobi_lh}
  \Description{A graph with twin y-axes: on the left is training loss history and on the right is Jacobi weight.  These values are evolving with training epoch (x-axis).}
\end{figure}

\begin{figure}
  \centering
  \includegraphics{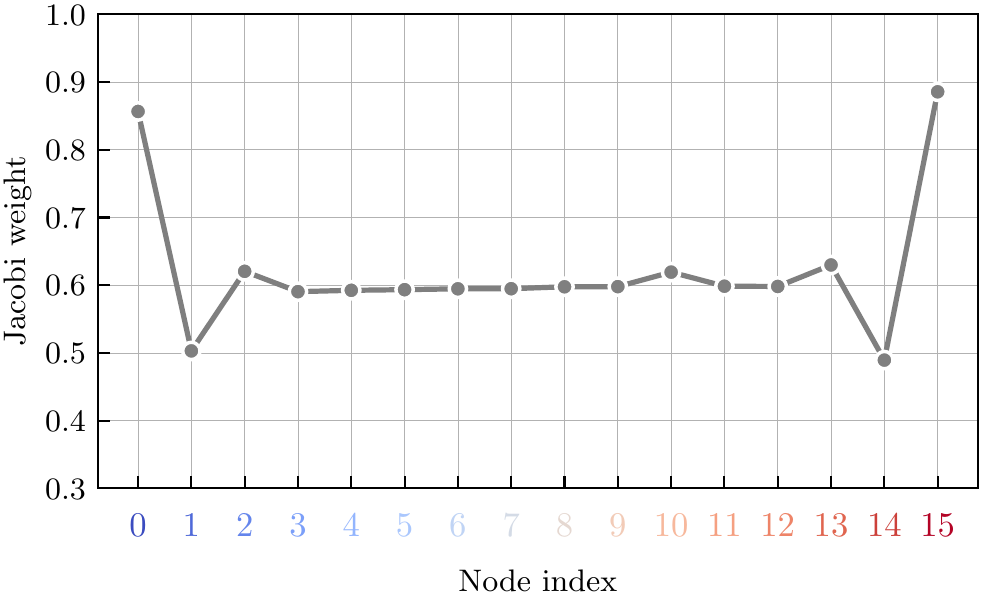}
  \caption{Final Jacobi weight values at each node; node indices are ordered by their $x$ position in the domain.}\label{fig:jacobi_omega}
  \Description{Final Jacobi weight of each node displayed spatially.  The two nodes at the ends of the domain are weighted higher than the interior nodes.}
\end{figure}

Training loss history and final values of the weights are shown in
\cref{fig:jacobi_lh,fig:jacobi_omega}.  In \cref{fig:jacobi_omega}, we observe that
the nodal weights at the two ends of the domain are
maximized; for our problem setup this mimics the behavior of
values being propagated from the boundaries inwards.  In the interior of the domain, the weights are observed to be close to $\omega = \frac{2}{3}$, which minimizes $\|I-\omega\mat{D}^{-1}\mat{A}\|_A$ over scalar weights, $\omega$.
\subsection{Heavyball Iteration}\label{subsec:heavyball}
The heavyball iteration~\cite{polyak1964heavyball} is a two-step iterative method that uses the information from the last two iterates to generate the next approximation.  It can be viewed as a gradient descent with a momentum term.

To minimize a differentiable function $f(\vec{x})$, consider the update step
\begin{equation}\label{eqn:polyak_general}
  \vec{x}^{(k+1)} = \vec{x}^{(k)} - \alpha \grad_{\vec{x}}f\big(\vec{x}^{(k)}\big) + \beta\big(\vec{x}^{(k)} - \vec{x}^{(k-1)}\big),
\end{equation}
for scalars $\alpha, \beta \in \mathbb{R}$.  Then, to solve systems of the form in \Cref{eqn:lin_solve_Axb} for fixed symmetric and positive-definite matrix $\mat{A}$ and vector $\vec{b}$, we define $f(\vec{x})$ as
\begin{equation}
  f(\vec{x}) = \frac{1}{2}\vec{x}^T\mat{A}\vec{x} - \vec{b}^T\vec{x}.
\end{equation}
Taking the gradient of $f$ with respect to $\vec{x}$ yields
\begin{equation}
  \grad_{\vec{x}}f(\vec{x}) = \mat{A}\vec{x} - \vec{b},
\end{equation}
where we have that if $\grad_{\vec{x}}f=\vec{0}$ then $\mat{x}$ is a solution to the matrix system $\mat{Ax}=\vec{b}$ that minimizes $f(\vec{x})$.

Similarly, we consider the example problem in \cref{subsec:jacobi} by generating $\mat{A}_N$ from~\Cref{eqn:mat_1d_fd} for $N=16$ and minimizing
\begin{equation}
  \ell = \sum_{j=1}^{J} \big(\mat{h}_{t}(\vec{x}_j, \vec{0};\:\alpha, \beta)\big)^T\mat{A}\mat{h}_{t}(\vec{x}_j, \vec{0};\:\alpha, \beta),
\end{equation}
where $\vec{h}_t(\vec{x}_j, \vec{b};\: \alpha, \beta)$ is the application of $t$ rounds of the heavyball iteration to $\vec{x}_j$ with right-hand-side $\vec{b}$ and parameters $\alpha$, $\beta$; here, we use $t=\frac{3}{4}N$, a fraction of the matrix size, as the number of iterations to optimize over.  Again, $\{\vec{x}_j\}$ is a set of $J$ random vectors in $\mathbb{R}^N$ with unit-normal entries.

\begin{figure}
  \centering
  \includegraphics{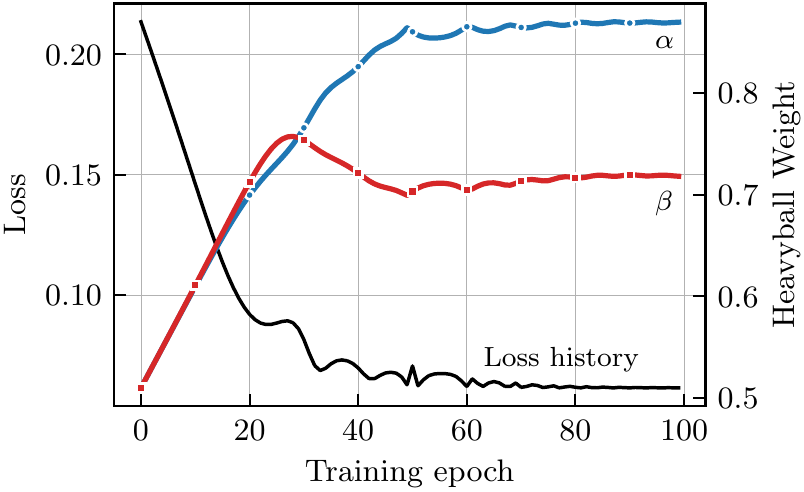}
  \caption{Parameter history for each heavyball training iteration, plotted along with the testing loss.}\label{fig:heavyball_param}
  \Description{A graph with twin y-axes: on the left is training loss history and on the right are the two heavyball weights.  These values are evolving with training epoch (x-axis).}
\end{figure}

We optimize the error after $\frac{3}{4}N$ iterations of the heavyball method, where $N$ is the number of unknowns in the system.  This fraction was chosen empirically: too few iterations leads to a divergent method while too many does not give any meaningful improvement.  The parameter history as a function of training epoch is shown in \cref{fig:heavyball_param}.

\begin{figure*}
  \centering
  \includegraphics{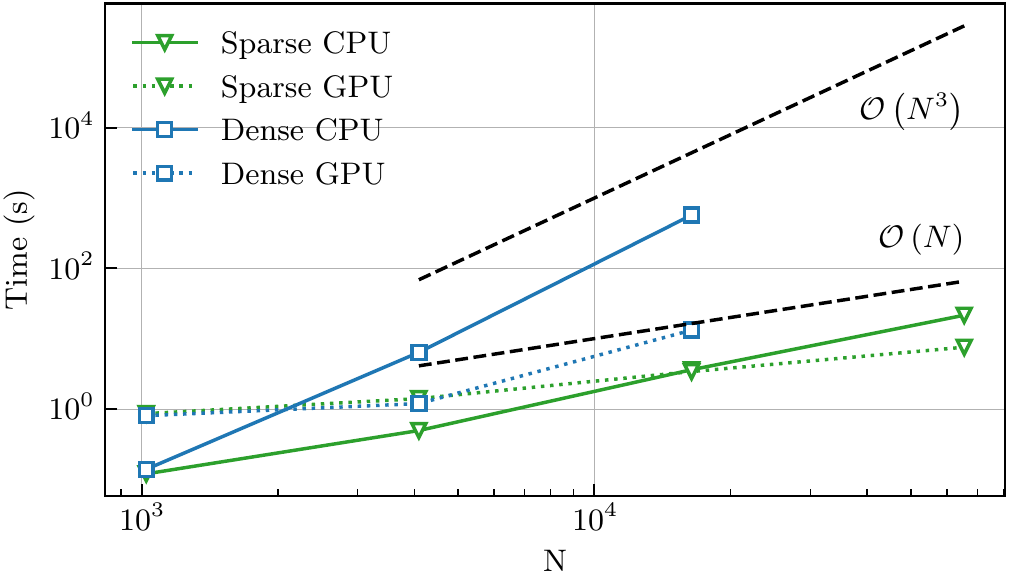}
  \caption{Timing result for optimizing over the heavyball iterations.  The sparse implementation shows approximately linear scaling with the problem size.}\label{fig:timing_results_heavyball}
\end{figure*}

We show timings of one training epoch as a function of the problem size in \cref{fig:timing_results_heavyball}.  Our sparse implementation has roughly linear scaling, while the dense implementation shows cubic scaling.
\subsection{Conjugate Gradient}\label{subsec:cg}
The conjugate gradient (CG) method is another iterative solver for matrix
equations of the form in~\Cref{eqn:lin_solve_Axb} when $\mat{A}$ is symmetric and positive definite (SPD\@), meaning that it is symmetric and has strictly positive eigenvalues. While it typically exhibits
faster convergence than both the Jacobi and heavyball methods,
convergence can be accelerated further by passing an approximate inverse to
$\mat{A}$, denoted by $\mat{M} \approx \mat{A}^{-1}$, with the assumption that
$\mat{M}$ is itself SPD, using the preconditioned conjugate gradient (PCG)
method~\cite{saad2003iterative}.  Finding a ``good'' preconditioner is an open question, and there are many methods that can generate a preconditioning scheme based on a priori knowledge
of the problem itself.  Here, we will find a preconditioner automatically via sparse optimization.

For our test problem, we use the 2D Poisson problem defined in \Cref{eqn:mat_2d_fd} with dimensions $N_x=N_y=8$, resulting in a matrix $\mat{A} \in \mathbb{R}^{64 \times 64}$.
We now aim to construct a matrix $\mat{M}$ that suitably approximates the inverse of $\mat{A}$.  We note that PCG requires an SPD preconditioner; to enforce this, we instead directly learn a lower triangular matrix, $\mat{L}$, and form $\mat{M}$ as
\begin{equation}
  \mat{M} = \mat{L}\mat{L}^T.
\end{equation}
This is similar to what is done in factored SPAI algorithms~\cite{KolotilinaFSAITheory1993,HuckleFSAI2003} for computing the approximate inverse to a sparse, SPD matrix.  We do not directly constrain $\mat{M}$ itself to be positive definite; indefinite or semidefinite preconditioners are unlikely to converge well and thus such a preconditioner will be avoided during optimization.  This optimization of a lower-triangular factorization for preconditioning is also similar to the method explored in~\cite{hausner2023neural}.

To find the entries in $\mat{M}$, we introduce the weighted loss
\begin{equation}
  \ell = \sum_{i=1}^{N_{\text{it}}} \Bigg(\frac{\gamma^{N_{\text{it}} - i}}{\sum_{j=1}^{N_{\text{it}}} \gamma^{N_{\text{it}}-j}}\Bigg) \frac{\big\| \vec{r}^{(i)} \big\|}{\|\vec{b}\|},
\end{equation}
where $N_{\text{it}}$ is the number of PCG iterations run (we use
$N_{\text{it}}=4$), $\vec{r}^{(i)}$ is the $i$th residual of the iteration,
$\vec{b}$ is the right-hand-side, and $\gamma\in(0,1]$ is a scaling  %
constant (we use $\gamma=0.6$).  With this form, we minimize the overall
weighted sum of the residual history, with later iterates being weighted more
than earlier ones. Experimentally, we observe improved
convergence of the optimization problem in comparison to minimizing only over the last
iterate.  This also avoids the problem of vanishing gradients as the number of
iterations is increased, as gradient information is used from each intermediate
iterate.

\begin{figure}
  \centering
  \includegraphics{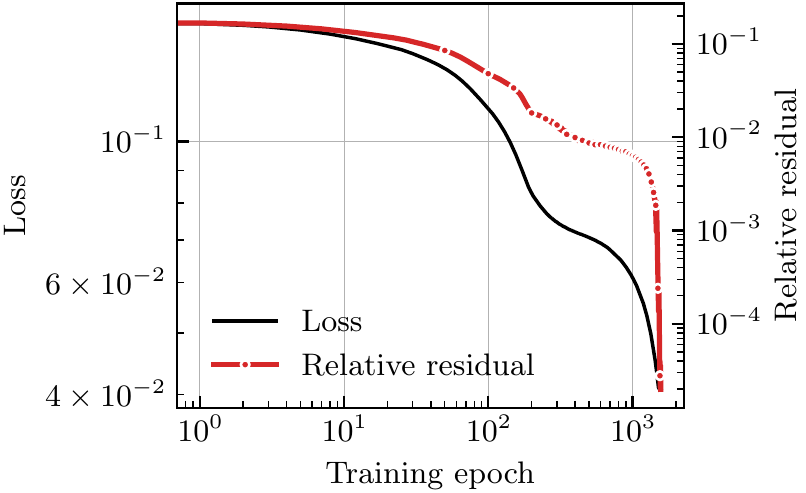}
  \caption{Loss history in constructing $\mat{M}$, the CG preconditioner.  The black line (left axis) is the loss and the red dashed line (right axis) is the relative residual after 4 PCG iterations.}\label{fig:pcg_lh}
  \Description{A graph with twin y-axes: on the left is training loss history and on the right is the relative residual of the solver at that point in training.  These values are evolving with training epoch (x-axis).}
\end{figure}

\begin{figure}
  \centering
  \includegraphics{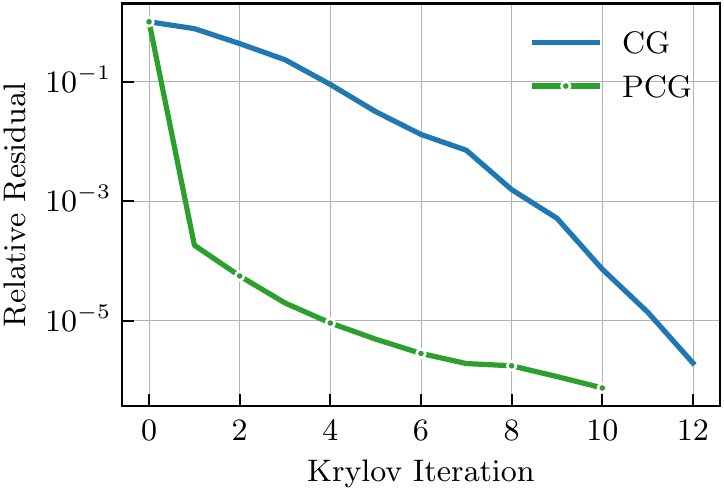}
  \caption{Relative residual history of a regular (non-preconditioned) CG method and the preconditioner found by optimization.}\label{fig:pcg_residual}
  \Description{A graph showing the relative residual, per Krylov iteration, of both the original and learned methods.  The learned method shows faster convergence.}
\end{figure}

We run the optimization over a lower bidiagonal $\mat{L}$, which gives
a tridiagonal $\mat{M}=\mat{L}\mat{L}^T$. The loss and final relative residual norm, $\|\vec{r}^{(4)}\|/\|\vec{b}\|$,
obtained during each training iteration are displayed in \cref{fig:pcg_lh}.
After optimization is complete, we obtain a matrix $\mat{M}$ that resembles a scaled, sparse approximate inverse to $\mat{A}$ with a tridiagonal sparsity pattern (cf.\ \cref{subsec:spai}).  The optimized preconditioner does indeed substantially improve the convergence rate, as shown in \cref{fig:pcg_residual} comparing the residual history with and without the preconditioner. Both solvers are run until a relative residual of $10^{-6}$ is achieved.

\begin{figure*}
  \centering
  \includegraphics{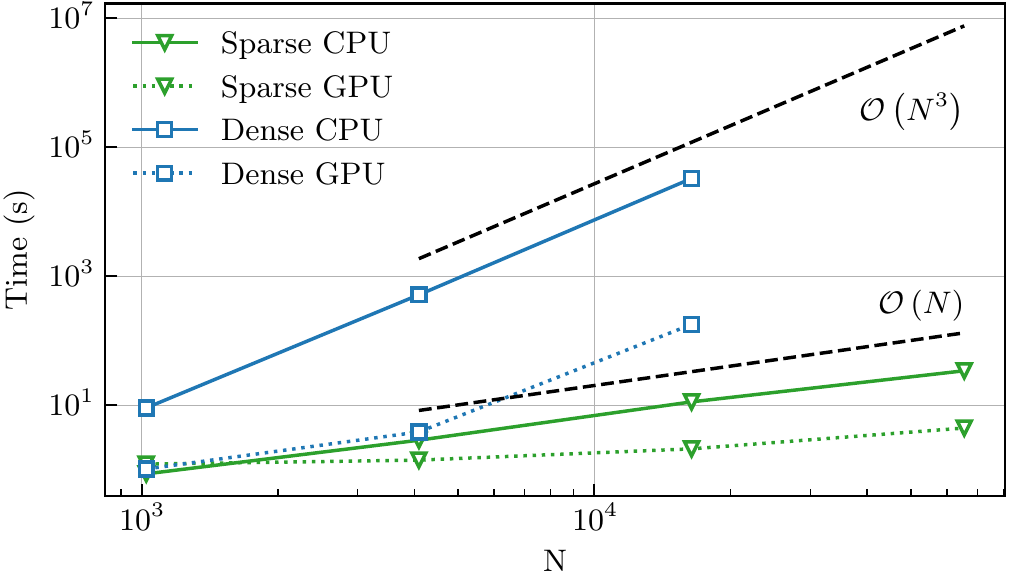}
  \caption{Timing result for training the preconditioned conjugate gradient method.  The sparse implementation shows linear scaling with the problem size.}\label{fig:timing_results_pcg}
\end{figure*}

Timing results of one training epoch versus problem size can be seen in \cref{fig:timing_results_pcg}.  The sparse implementation shows very clear linear scaling as the problem size increases, while the dense implementation shows cubic scaling because of the $\mat{M}=\mat{L}\mat{L}^T$ product that is necessary in order to form the preconditioner.
\subsection{Graph Neural Networks}\label{subsec:gcn}
Graph neural networks have become massively popular in recent years due to their ability to perform general inferencing tasks on semi-structured data~\cite{Wu2021GraphnetSurvey}.  Most graph network implementations in large libraries such PyG~\cite{pyg} or DGL~\cite{dgl} have their own bespoke data structures and software implementations; however, we will show here that spectral graph convolutions~\cite{BrunaSpectral2013} can be performed in a simple and general way using our sparse kernels.

The GCN layer~\cite{kipf2017gcn} is a graph convolutional layer that
\textit{convolves} node features on a graph.
Let the (weighted) adjacency matrix of a graph be given by $\mat{A} \in
\mathbb{R}^{N \times N}$, and denote $\mat{X}^{(i)} \in \mathbb{R}^{N \times C}$ as the
$C$-dimensional node features at layer $i$. The convolution of node features
is then represented as
\begin{equation}
  \mat{X}^{(i+1)} = \tilde{\mat{D}}^{-\frac{1}{2}} \tilde{\mat{A}} \tilde{\mat{D}}^{-\frac{1}{2}} \mat{X}^{(i)} \mat{\Theta}^{(i+1)}, \label{eqn:gcn_update}
\end{equation}
where $\mat{\Theta}^{(i+1)} \in \mathbb{R}^{C \times F}$ is the weight matrix for
layer $i+1$, $\tilde{\mat{A}} = \mat{A} + \mat{I}$, and $\tilde{\mat{D}}$ is the
diagonal matrix extracted from $\tilde{\mat{A}}$.

\begin{figure}
  \centering
  \begin{tikzpicture}[remember picture, overlay]
\coordinate (x_shift) at ([xshift=\linewidth]pic cs:gcn_rowsum_top);
\coordinate (gcn_rowsum_top_rect) at ({pic cs:gcn_rowsum_mid} -| x_shift);
\coordinate (gcn_rowsum_mid_rect) at ({pic cs:gcn_rowsum_bot} -| x_shift);
\path[overlay, fill=red!10] ($(pic cs:gcn_rowsum_top) + (-0.1,0.23)$) rectangle ($(gcn_rowsum_top_rect) + (-2.5,-0.05)$);
\path[overlay, fill=green!10] ($(pic cs:gcn_rowsum_mid) + (-0.1,0.20)$) rectangle ($(gcn_rowsum_mid_rect) + (-2.5,0.20)$);
\end{tikzpicture}
\begin{minted}[xleftmargin=4em,escapeinside=||]{Python}
  import torch
  import torch.nn
  import numml.sparse as sp

  class GCNConv(torch.nn.Module):
      def __init__(self, in_channels, out_channels):
          super().__init__()

          self.weights = torch.nn.Parameter(
              torch.randn(in_channels, out_channels))
          self.bias = torch.nn.Parameter(torch.randn(out_channels))

      def forward(self, graph, X):
|\tikzmark{gcn_rowsum_top}|-         D = (graph.sum(dim=1) + 1.) ** -0.5           # Pre-compute D
|\tikzmark{gcn_rowsum_mid}|+         D = (graph.row_sum() + 1.) ** -0.5            # Pre-compute D
|\tikzmark{gcn_rowsum_bot}|          XTheta = X @ self.weights                     # X * Theta
          DXTheta = (D[:, None] * XTheta)               # D * X * Theta
          C = D[:, None] * (graph @ DXTheta + DXTheta)  # (eqn |\ref{eqn:gcn_update}|)

          return C + self.bias
\end{minted}
  \caption{Implementation code for the graph convolutional layer (GCN).  Differences between the sparse (green, +) and dense (red, -) implementations are shown with highlighted lines.}\label{fig:snippet_gcn}
\end{figure}

In this example, we expose the GCN layer to our optimized underlying sparse-matrix operations. Depending on the order of operations, computing the output in \Cref{eqn:gcn_update} can be viewed as a series of sparse-dense matrix multiplies (multiplying from right-to-left), or as two sparse-sparse multiplies followed by dense multiplies.  In the following, we consider the former method.  Our reference implementation can be seen in \cref{fig:snippet_gcn}.

\begin{figure}
  \centering
  \includegraphics{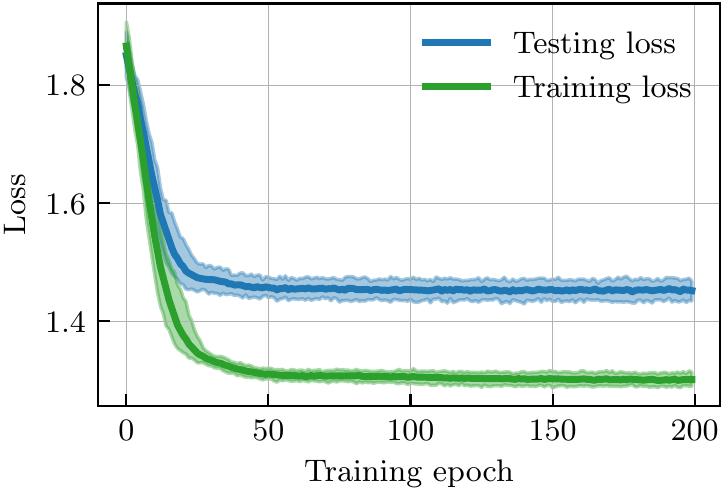}
  \caption{Loss history for training the GNN over the CiteSeer dataset.
    Lines denote the mean (over 100 runs) while shaded regions are two standard deviations from mean.}\label{fig:gnn_lh}
  \Description{A figure showing the training history of the GNN on the training and testing split of the CiteSeer dataset.
    Both converge in approximately 50 iterations.  The training loss is smaller overall.}
\end{figure}

Using a GCN layer, we consider the semi-supervised CiteSeer example from~\cite{kipf2017gcn}. This dataset has \num{3327} nodes, \num{4732} edges, and six classes. We train a two-layer GCN with ReLU and sigmoid activations after the first and second layers, respectively, and use a cross-entropy loss to optimize the predicted labels on each node, which follows the same setup as in~\cite{kipf2017gcn}. Likewise, between each GCN layer is a dropout layer with $p=0.5$, and we train with an Adam optimizer using a learning rate of $0.01$, $L_2$ regularization of \num{5e-4}, and $16$-dimensional representations for the hidden node features.  This is trained for \num{200} epochs over \num{100} random initializations and training history can be seen in \cref{fig:gnn_lh}.

\begin{figure}
  \centering
  \includegraphics{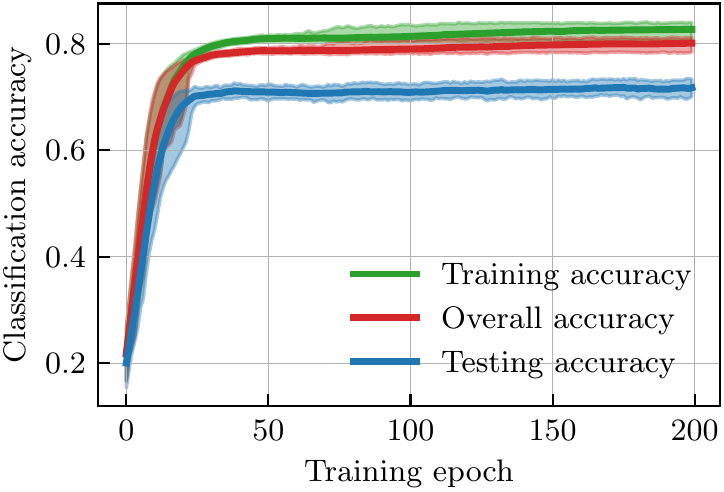}
  \caption{Classification accuracy of the networks per training epoch.
    Lines denote the mean (over 100 runs) while shaded regions are two standard deviations from mean.}\label{fig:gnn_accuracy}
  \Description{A figure showing the overall, training, and testing accuracy of the GNN versus training epoch.
    The training converges in approximately 50 iterations.}
\end{figure}

In~\cref{fig:gnn_accuracy}, we observe comparable accuracy to the roughly 70\% classification accuracy on the CiteSeer dataset.
To benchmark the performance of the underlying sparse methods, we train the graph neural network on four different datasets: the Cora ($N=\num{2708}$), CiteSeer ($N=\num{3327}$),
and PubMed ($N=\num{19717}$) datasets from~\cite{kipf2017gcn,CiteSeerDataset}; and the Flickr ($N=\num{89250}$) dataset from~\cite{FlickrDataset}.
Additionally, we include several larger graphs from the Suitesparse matrix collection~\cite{suitesparse}; these graphs range in size from $\num{100000}$ nodes
to $\num{914231}$ nodes, and have a roughly constant number of edges to nodes (each node has on average $5$ incident edges).

\begin{figure}
  \centering
  \includegraphics{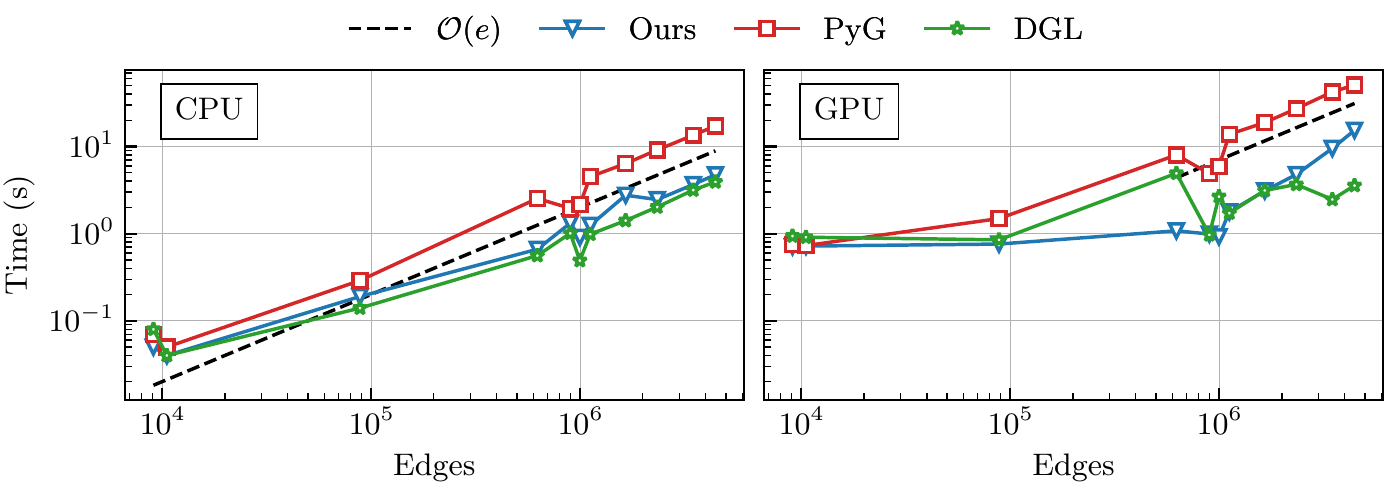}
  \caption{Scalings with number of edges for the GCN implementations on CPU and GPU.}\label{fig:gcn_scalings}
  \Description{GCN scalings on both CPU and GPU, compared to the number of edges in each graph.}
\end{figure}

We compare our running times with the existing PyG and DGL graph network libraries in \cref{tab:gcn_timings} and \cref{fig:gcn_scalings}.
In~\cref{tab:gcn_timings}, $N$ refers to the number of nodes present in each graph; in the figure, however, we plot times vs. number of edges, noting that we have roughly a constant proportion of edges to nodes.
The CPU implementation of our method shows linear scaling, comparable to PyG and DGL, and all three methods achieve similar running times.
When our method is run on GPU, we also achieve similar running times to PyG and DGL, though a slight inflection (faster than linear run-time)
can be seen towards the larger problem sizes; a consequence of using the CSR matrix format is that a matrix transpose is needed to compute the VJP in
the backward pass.  Because PyG and DGL use coordinate-based COO formats, they see slower asymptotic growth as the problem size increases.
Overall, we are able to achieve impressive performance using our off-the-shelf kernels to perform spectral graph convolution in comparison to highly tuned libraries for this task.

\begin{table}
  \centering
  \sisetup{input-ignore={,},input-decimal-markers={.},group-separator={,}}
\begin{tabular}{
@{}
l %
|
l %
|
S[table-auto-round, table-format=4.2]
S[table-auto-round, table-format=4.2]
S[table-auto-round, table-format=4.2]
S[table-auto-round, table-format=4.2]
S[table-auto-round, table-format=4.2]
S[table-auto-round, table-format=4.2]
@{}}
\toprule
{\footnotesize Implementation} & {\footnotesize Device} & {\footnotesize $N=\num{2708}$} & {\footnotesize $N=\num{3327}$} & {\footnotesize $N=\num{19717}$} & {\footnotesize $N=\num{89250}$} & {\footnotesize $N=\num{133769}$} & {\footnotesize $N=\num{343565}$} \\
\midrule
{\multirow{2}{*}{Sparse (\textbf{ours})}}
& {CPU} & 0.035 & 0.052 & 0.193 & 1.321 & 0.66243 & 2.76251 \\
& {GPU} & 0.723 & 0.708 & 0.758 & 0.987 & 1.07845 & 3.09787 \\
\midrule
{\multirow{2}{*}{Sparse (DGL)}}
& {CPU} & 0.037 & 0.076 & 0.137 & 1.011 & 0.56063 & 1.4135 \\
& {GPU} & 0.910 & 0.940 & 0.854 & 0.968 & 4.9281 & 3.12908 \\
\midrule
{\multirow{2}{*}{Sparse (PyG)}}
& {CPU} & 0.049 & 0.071 & 0.294 & 1.945 & 0.56063 & 6.37213 \\
& {GPU} & 0.746 & 0.734 & 1.488 & 4.880 & 8.00548 & 18.83343 \\
\midrule
{\multirow{2}{*}{Dense (Torch)}}
& {CPU} & 0.131 & 0.157 & 3.893 & \multicolumn{1}{r}{--} & \multicolumn{1}{r}{--} & \multicolumn{1}{r}{--} \\
& {GPU} & 0.748 & 0.701 & 0.784 & \multicolumn{1}{r}{--} & \multicolumn{1}{r}{--} & \multicolumn{1}{r}{--} \\
\bottomrule
\end{tabular}

  \caption{A subset of timings for training the graph neural network examples for one training epoch, averaged over 5 runs and compared against graph network libraries PyG~\cite{pyg} and DGL~\cite{dgl}.  Entries with (--) indicate runs that did not complete due to insufficient memory needed for both forward and back-propagation.  Units are in seconds.}\label{tab:gcn_timings}
\end{table}

\subsection{Optimized Domain Decomposition}\label{subsec:dd}
Domain decomposition methods are often highly effective in computing the numerical solution to partial differential equations; however, their setup and optimal parameter selection requires careful analysis that is infeasible for complex problems, such as those on unstructured domains.  In the work by \citet{taghibakhshi2022mloras}, the authors develop a learned domain decomposition algorithm that performs the setup in an automatic way, alleviating the need for tedious parameter selection or analysis.  In this section, we will describe how the use of our sparse kernels can be used to speed up training, and also greatly enhance the scalability of the method itself.

Formally, the domain decomposition algorithm takes as input some sparse matrix $\mat{A} \in \mathbb{R}^{N \times N}$ and a partitioning of the index set of nodes $\set{D} = \big\{1, \ldots, N\}$, into $S$ disjoint subdomains $\set{D}_1, \set{D}_2, \ldots, \set{D}_S$ such that $\set{D}_i \ \cap\ \set{D}_j = \emptyset \enskip \forall i,j$ where $i \neq j$, and $\bigcup_i \set{D}_i = \set{D}$.  We denote piecewise-constant restriction operators from $\set{D}$ to subdomain $\set{D}_i$ by $\mat{R}^{0}_i$.  To allow for some overlap between domains, additionally let $\mat{R}^{\delta}_i$ be the restriction to $\set{D}_i^{\delta}$, the union of $\set{D}_i$ \textit{and} the set of nodes that have at most distance $\delta$ from its boundary.  Note, however, that $\mat{R}^{0}_{i}$ and $\mat{R}^{\delta}_i$ may introduce different orderings to nodes on the local subdomain $\set{D}_i$, thus we also define $\tilde{\mat{R}}^{\delta}_i$ to have the same shape and row ordering as in $\mat{R}^{\delta}_i$ but with nonzero rows only for nodes in $\set{D}_i$ itself: extended node values are masked off.

Following the standard RAS domain decomposition
approach~\cite{Toselli2005}, we then form an approximate inverse or preconditioner to $\mat{A}$ as
\begin{equation}
  \mat{A}^{-1} \approx \mat{M}_{\text{DD}} = \sum_{i=1}^{S} \Big(\tilde{\mat{R}}^{\delta}_i\Big)^T \mat{A}_i^{-1} \mat{R}^{\delta}_i,
\end{equation}
where $\mat{A}_i$ is a projection (or re-discretization) of the full problem to subdomain $\set{D}_i^{\delta}$.

\begin{figure}
  \centering
  \includegraphics{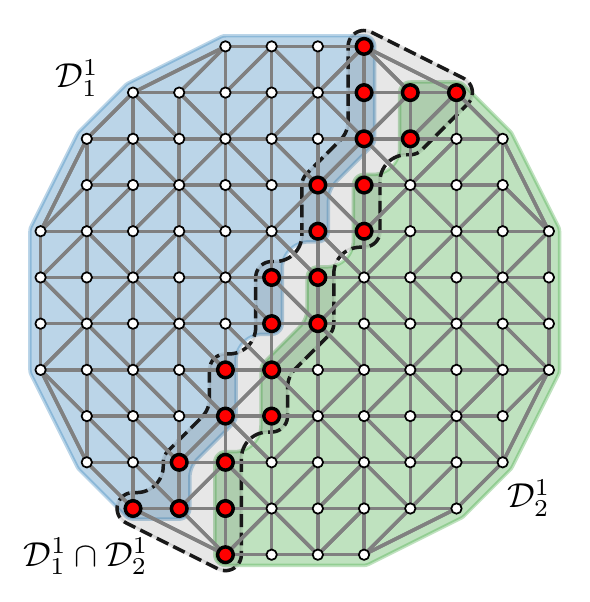}
  \caption{An example circular domain split into two overlapping sets, $\set{D}_1^1$ and $\set{D}_2^1$.  Transmission conditions are learned on edges in their overlap, denoted by the region with a dashed outline and larger red points.}\label{fig:oras_agg}
  \Description{A meshed, circular domain is split into two even subdomains.  Nodes that stride this split boundary are denoted in red.
    Edges between these nodes are highlighted; these edges are the ones learned by the method.}
\end{figure}

An improved approximation can be obtained if a modified form is used~---~that is,
\begin{equation}
  \tilde{\mat{A}_i} = \mat{A}_i + \mat{L}_i,
\end{equation}
where $\mat{L}_i$ is some learned matrix containing entries only on the subdomain \textit{boundary}; we are, in essence,
learning the interface or transmission conditions for each subdomain, see \cref{fig:oras_agg}.
To output $\mat{L}_i$, we first generate a set of node values, $\vec{d}$, defined as
\begin{equation}
  d_i =
  \begin{cases}
    1 & \text{node } i \text{ is along a subdomain boundary,} \\
    0 & \text{otherwise.}
  \end{cases}
\end{equation}
These are then passed to a graph neural network, along with the matrix $\mat{A}$, through several node and edge convolutions (see Appendix C from~\cite{taghibakhshi2022mloras} for exact architecture) with learnable parameters $\theta$ to output a new matrix $\hat{\mat{L}}^{(\theta)}$, such that
$\mask(\hat{\mat{L}}^{(\theta)}) = \mask(\mat{A})$.  For each subdomain $\set{D}_i$, we then mask $\hat{\mat{L}}^{(\theta)}$ so that $\mat{L}_i^{(\theta)}$
contains nonzero entries only between the boundary nodes in $\set{D}_i$.  This gives us the learned preconditioner
\begin{equation}
  \mat{M}^{(\theta)} = \sum_{i=1}^S \Big(\tilde{\mat{R}}_i^{\delta}\Big)^T \Big(\mat{R}_i^{\delta}\mat{A}\Big(\mat{R}_i^{\delta}\Big)^T + \mat{L}_i^{(\theta)}\Big)^{-1} \mat{R}_i^{\delta}. \label{eqn:m_theta}
\end{equation}

To optimize the network parameters to output optimal interface conditions, we define the error-propagation operator
\begin{equation}
  \mat{T}^{(\theta)} = \mat{I} - \mat{M}^{(\theta)} \mat{A},
\end{equation}
as the map of the error over each iteration of the domain decomposition solver.
An obvious choice for a loss is to minimize the spectral norm of $\mat{T}^{(\theta)}$, as this directly measures how fast the solver converges in the worst case.  Computing this can be difficult in practice, however.  To avoid expensive eigendecompositions that may particularly cause trouble with gradient propagation, we instead use a stochastic approximation:  let $\set{X} = \{\vec{x}_1, \vec{x}_2, \ldots, \vec{x}_J\}$ be a set of $J$ unit vectors in $\mathbb{R}^N$ whose entries are uniformly distributed; we approximate the loss by a stochastic relaxation of the induced matrix norm,
\begin{equation}
  \ell \approx \max_{\vec{x} \in \set{X}} \Big\|\Big(\mat{T}^{(\theta)}\Big)^k \vec{x}\Big\|_2,
\end{equation}
where $k$ is the number of solver iterations using for the training.

We, thus, have a loss that we can use to train the domain decomposition method in an end-to-end fashion.
The methods in~\cite{taghibakhshi2022mloras} are constrained to training on small problem sizes, as their automatic differentiation package does not support the sparse-sparse product $\mat{M}^{(\theta)} \mat{A}$; both $\mat{M}^{(\theta)}$ and $\mat{A}$ are stored in their dense representation for training the GNN\@.  With the differentiable sparse kernels that we have introduced in \cref{sec:spops}, we reimplement their training routine using sparse operations where applicable.

\begin{figure*}
  \centering
  \includegraphics{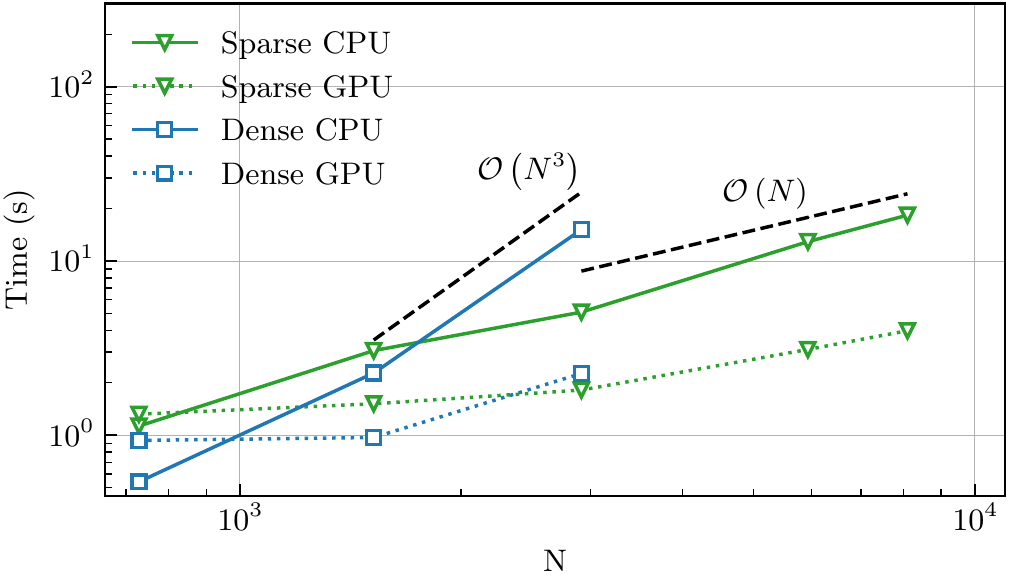}
  \caption{Timing result for one training epoch of the learned domain decomposition solver.  The sparse implementation has approximately linear scaling with the problem size, while the dense implementation shows much faster growth.}\label{fig:timing_results_oras}
\end{figure*}

The results shown in~\cref{fig:timing_results_oras} for the domain decomposition method are particularly striking, as the existing dense implementation is unable to scale larger than problems of size $N=\num{5929}$ without running out of available memory, whereas the sparse implementation continues to run for larger problems with roughly linear time complexity with respect to the problem size.
\subsection{Sparse Approximate Inverses}\label{subsec:spai}
Sparse approximate inverses (SPAI)~\cite{grotehuckle1997spai,HuckleFSAI2003,BrokerSPAIRelaxation2002,KolotilinaFSAITheory1993,BrokerThesis} are a family of methods for finding explicit approximate inverses to sparse linear systems.  These are often used for preconditioning, such as in CG, or as relaxation schemes in multigrid solvers.  The basic SPAI algorithm minimizes the function
\begin{equation}
  \ell = \|\mat{I} - \mat{M}\mat{A}\|^2_F, \label{eqn:spai_loss}
\end{equation}
where $\mat{A} \in \mathbb{R}^{N \times N}$ is some sparse system and $\mat{M}\in\mathbb{R}^{N \times N}$ is the approximate sparse inverse such that $\mat{M}\mat{A} \approx \mat{I}$.  Normally, the loss in \Cref{eqn:spai_loss} is decomposed into parallel least squares problems, as we have
\begin{equation}
  \|\mat{I} - \mat{M}\mat{A}\|^2_F = \sum_{i=1}^N \big\|(\mat{I} - \mat{M}\mat{A})\vec{e}_i\big\|_2^2,
\end{equation}
with $\vec{e}_i$ denoting the $i^{\text{th}}$ canonical unit vector,
which can be further rearranged into $n$ minimization problems over the rows of $\mat{M}$.  However, we show that it is feasible to minimize \Cref{eqn:spai_loss} directly with respect to the
nonzero entries of $\mat{M}$.

In some variations of the SPAI algorithm, the sparsity pattern of the approximate inverse $\mat{M}$ is allowed to be chosen dynamically according to some tolerance~\cite{BrokerThesis}: fill-in is introduced into the approximate inverse during iterations to reduce row-wise residual.  In our case, we force the sparsity pattern of $\mat{M}$ to remain static.  Additionally, $\mask(\mat{M}) = \mask(\mat{A})$, meaning the inverse has the same sparsity as the system itself.
\begin{figure}
  \centering
  \includegraphics{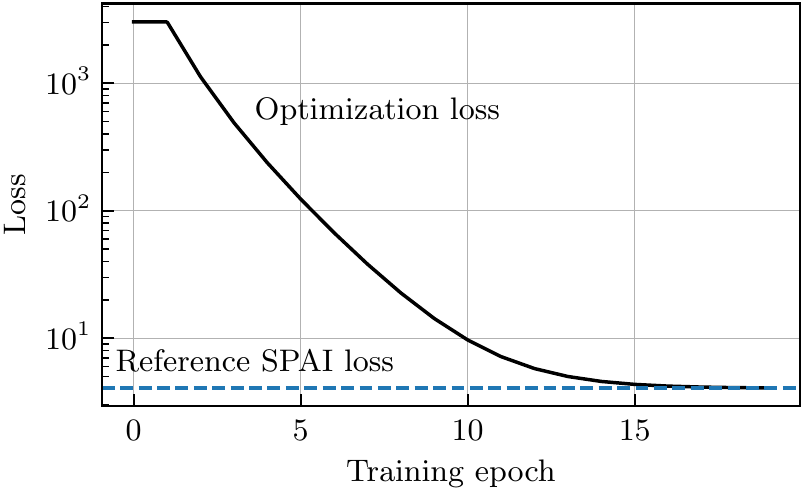}
  \caption{Loss history obtained by directly optimizing the SPAI objective function (black), as compared to the loss of the output from the traditional SPAI algorithm as formulated in~\cite{grotehuckle1997spai} (blue).  Both methods converge to roughly the same result.}\label{fig:spai_lh}
  \Description{A figure showing the optimization loss curve compared to the reference SPAI loss as a horizontal line.  The loss curve approaches the regular loss after approximately 20 iterations.}
\end{figure}

\begin{figure*}
  \centering
  \includegraphics{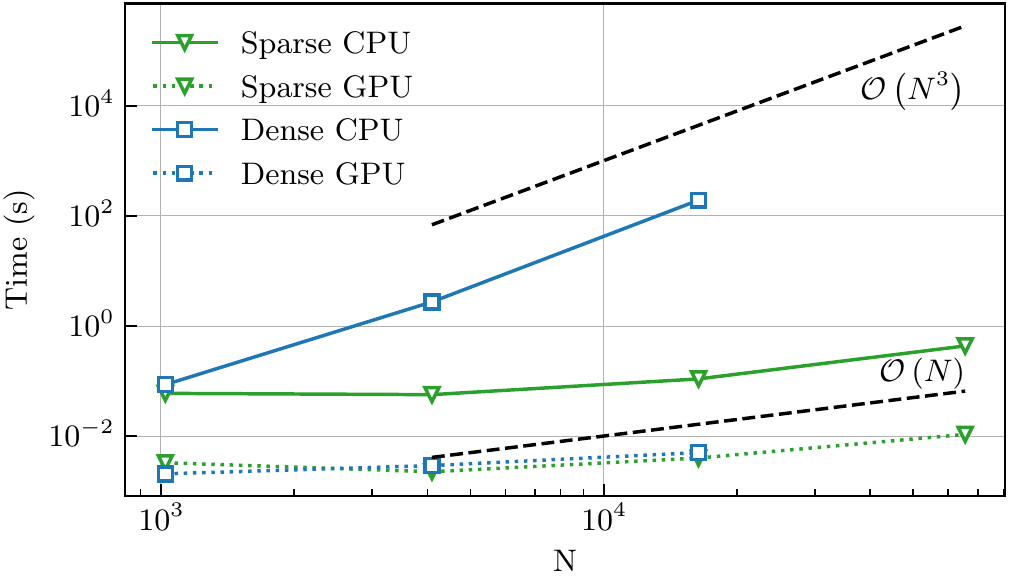}
  \caption{Timing result for the SPAI application.  The sparse implementation has a linear $\mathcal{O}(n)$ scaling, while the dense implementation shows a asymptotic higher complexity.}\label{fig:timing_results_spai}
  \Description{Timing plots for the Sparse approximate inverse implementation.}
\end{figure*}

We use the two-dimensional Poisson test problem as defined in \Cref{eqn:mat_2d_fd}, as this gives a more complex sparsity problem for the optimization routine to find an inverse over.  For our first results, we take $N_x=N_y=8$, resulting in a system with shape $\mat{A} \in \mathbb{R}^{64 \times 64}$, then consider scaling with $N = N_x\times N_y$ in \cref{fig:timing_results_spai}.  To perform the optimization itself, $\mat{M}$ is initialized to $\mat{M} = \mask(\mat{A})$, meaning that all respective nonzero entries in $\mat{A}$ are initialized to $1$ in $\mat{M}$.  Gradient descent is then run over the nonzero entries of $\mat{M}$ until the gradient norm of \Cref{eqn:spai_loss}, computed by automatic differentiation, is below $0.01$.

The result of optimizing the sparse inverse over the test problem is shown in \cref{fig:spai_lh}.  Our optimization-based method is compared to a Python implementation of the SPAI algorithm defined in~\cite{grotehuckle1997spai}; for our small problem, we achieve roughly the same result in approximately 20 iterations of gradient descent.  An interesting side effect of using our gradient kernels to perform the optimization is that we have a GPU implementation ``for free'': previous works porting SPAI to GPU~\cite{WangCUDASPAI2021,BertacciniGPUSPAI2016} have required special considerations to effectively exploit massively parallel architectures.

Using this optimization method, we see roughly linear scaling of our sparse method in \cref{fig:timing_results_spai}, as opposed to cubic scaling in the dense case.  However, it is interesting to note that that the GPU versions show similar timings no matter if the underlying implementation is dense or sparse; the high amount of parallelism in the dense case will counteract any extra work being done. However, all dense implementations are unable to run at all for the case of $N=\num{65536}$, as too much memory is consumed storing the matrix representation and intermediate data for backpropagation.
\section{Conclusions}
In this work, we describe the implementation of a framework for
automatically computing the gradient of computations with sparse matrix
operations. We outline the backward propagation update rules and describe
how they can be efficiently implemented on CPUs and GPUs.  We
demonstrate a range of applications that can be
optimized using these sparse matrix primitives.  Finally, we present timing and
scaling results to show that our implementations are both scalable with respect
to dense linear algebra routines, and competitive in the forward pass with
other sparse linear algebra packages.

There are numerous directions of future work that we envision from this paper.
Perhaps the most straightforward is the implementation of more complex
sparse linear algebra operations, such as eigensolver routines, linear least
squares, etc.  The implementation of a full sparse direct solve routine on the GPU
could be immensely useful, and whose implementation could follow scalable approaches
such as in~\cite{gaihre2022gsofa}.

In this work, we choose to use CSR because it is well-used in traditional scientific computing
settings and provides good access to the matrix rows in memory.  However, CSR does not always
provide optimal access to the matrix elements in the backwards VJP passes.  Thus, an interesting
future work could be to look at different sparse matrix representations (or devise an entirely new
one) that provides balanced performance in both the forward and backwards passes for the different kernels.

Our reference PyTorch interface code providing the above sparse kernels as well as example applications can be found at \url{https://github.com/nicknytko/numml}.

\begin{acks}
  This research was enabled in part by support provided by ACENET (\url{www.ace-net.ca}), and the Digital Research Alliance of Canada (\url{alliancecan.ca}).  The work of SM was partially supported by an NSERC Discovery Grant.
\end{acks}
\bibliographystyle{ACM-Reference-Format}
\bibliography{paper_num_ml}


\begin{thebibliography}{53}


\ifx \showCODEN    \undefined \def \showCODEN     #1{\unskip}     \fi
\ifx \showDOI      \undefined \def \showDOI       #1{#1}\fi
\ifx \showISBNx    \undefined \def \showISBNx     #1{\unskip}     \fi
\ifx \showISBNxiii \undefined \def \showISBNxiii  #1{\unskip}     \fi
\ifx \showISSN     \undefined \def \showISSN      #1{\unskip}     \fi
\ifx \showLCCN     \undefined \def \showLCCN      #1{\unskip}     \fi
\ifx \shownote     \undefined \def \shownote      #1{#1}          \fi
\ifx \showarticletitle \undefined \def \showarticletitle #1{#1}   \fi
\ifx \showURL      \undefined \def \showURL       {\relax}        \fi
\providecommand\bibfield[2]{#2}
\providecommand\bibinfo[2]{#2}
\providecommand\natexlab[1]{#1}
\providecommand\showeprint[2][]{arXiv:#2}

\bibitem[Abadi et~al\mbox{.}(2016)]%
        {tensorflow}
\bibfield{author}{\bibinfo{person}{Mart{\'\i}n Abadi}, \bibinfo{person}{Paul
  Barham}, \bibinfo{person}{Jianmin Chen}, \bibinfo{person}{Zhifeng Chen},
  \bibinfo{person}{Andy Davis}, \bibinfo{person}{Jeffrey Dean},
  \bibinfo{person}{Matthieu Devin}, \bibinfo{person}{Sanjay Ghemawat},
  \bibinfo{person}{Geoffrey Irving}, \bibinfo{person}{Michael Isard},
  \bibinfo{person}{Manjunath Kudlur}, \bibinfo{person}{Josh Levenberg},
  \bibinfo{person}{Rajat Monga}, \bibinfo{person}{Sherry Moore},
  \bibinfo{person}{Derek~G. Murray}, \bibinfo{person}{Benoit Steiner},
  \bibinfo{person}{Paul Tucker}, \bibinfo{person}{Vijay Vasudevan},
  \bibinfo{person}{Pete Warden}, \bibinfo{person}{Martin Wicke},
  \bibinfo{person}{Yuan Yu}, {and} \bibinfo{person}{Xiaoqiang Zheng}.}
  \bibinfo{year}{2016}\natexlab{}.
\newblock \showarticletitle{{TensorFlow}: A System for {Large-Scale} Machine
  Learning}. In \bibinfo{booktitle}{\emph{12th USENIX Symposium on Operating
  Systems Design and Implementation (OSDI 16)}}. \bibinfo{publisher}{USENIX
  Association}, \bibinfo{address}{Savannah, GA}, \bibinfo{pages}{265--283}.
\newblock
\showISBNx{978-1-931971-33-1}
\urldef\tempurl%
\url{https://www.usenix.org/conference/osdi16/technical-sessions/presentation/abadi}
\showURL{%
\tempurl}


\bibitem[Baydin et~al\mbox{.}(2017)]%
        {baydinAutodiffsurvey2015}
\bibfield{author}{\bibinfo{person}{At\i{}l\i{}m~G\"{u}nes Baydin},
  \bibinfo{person}{Barak~A. Pearlmutter}, \bibinfo{person}{Alexey~Andreyevich
  Radul}, {and} \bibinfo{person}{Jeffrey~Mark Siskind}.}
  \bibinfo{year}{2017}\natexlab{}.
\newblock \showarticletitle{Automatic Differentiation in Machine Learning: A
  Survey}.
\newblock \bibinfo{journal}{\emph{J. Mach. Learn. Res.}} \bibinfo{volume}{18},
  \bibinfo{number}{1} (\bibinfo{date}{jan} \bibinfo{year}{2017}),
  \bibinfo{pages}{5595–5637}.
\newblock
\showISSN{1532-4435}


\bibitem[Bell et~al\mbox{.}(2012)]%
        {2012_BeDaOl_amggpu}
\bibfield{author}{\bibinfo{person}{N. Bell}, \bibinfo{person}{S. Dalton}, {and}
  \bibinfo{person}{L. Olson}.} \bibinfo{year}{2012}\natexlab{}.
\newblock \showarticletitle{Exposing Fine-Grained Parallelism in Algebraic
  Multigrid Methods}.
\newblock \bibinfo{journal}{\emph{SIAM Journal on Scientific Computing}}
  \bibinfo{volume}{34}, \bibinfo{number}{4} (\bibinfo{year}{2012}),
  \bibinfo{pages}{C123--C152}.
\newblock
\urldef\tempurl%
\url{https://doi.org/10.1137/110838844}
\showDOI{\tempurl}


\bibitem[Bertaccini and Filippone(2016)]%
        {BertacciniGPUSPAI2016}
\bibfield{author}{\bibinfo{person}{Daniele Bertaccini} {and}
  \bibinfo{person}{Salvatore Filippone}.} \bibinfo{year}{2016}\natexlab{}.
\newblock \showarticletitle{Sparse approximate inverse preconditioners on high
  performance {GPU} platforms}.
\newblock \bibinfo{journal}{\emph{Computers \& Mathematics with Applications}}
  \bibinfo{volume}{71} (\bibinfo{date}{2} \bibinfo{year}{2016}),
  \bibinfo{pages}{693--711}.
\newblock
Issue 3.
\showISSN{0898-1221}
\urldef\tempurl%
\url{https://doi.org/10.1016/J.CAMWA.2015.12.008}
\showDOI{\tempurl}


\bibitem[Bienz et~al\mbox{.}(2019)]%
        {2018_BiGrOl_nodeawarespmv}
\bibfield{author}{\bibinfo{person}{Amanda Bienz}, \bibinfo{person}{William~D.
  Gropp}, {and} \bibinfo{person}{Luke~N. Olson}.}
  \bibinfo{year}{2019}\natexlab{}.
\newblock \showarticletitle{Node aware sparse matrix–vector multiplication}.
\newblock \bibinfo{journal}{\emph{J. Parallel and Distrib. Comput.}}
  \bibinfo{volume}{130} (\bibinfo{year}{2019}), \bibinfo{pages}{166--178}.
\newblock
\showISSN{0743-7315}
\urldef\tempurl%
\url{https://doi.org/10.1016/j.jpdc.2019.03.016}
\showDOI{\tempurl}


\bibitem[Bonfatti et~al\mbox{.}(1974)]%
        {microwave1974}
\bibfield{author}{\bibinfo{person}{F. Bonfatti}, \bibinfo{person}{V.A. Monaco},
  {and} \bibinfo{person}{P. Tiberio}.} \bibinfo{year}{1974}\natexlab{}.
\newblock \showarticletitle{Microwave Circuit Analysis by Sparse-Matrix
  Techniques}.
\newblock \bibinfo{journal}{\emph{IEEE Transactions on Microwave Theory and
  Techniques}} \bibinfo{volume}{22}, \bibinfo{number}{3}
  (\bibinfo{year}{1974}), \bibinfo{pages}{264--269}.
\newblock
\urldef\tempurl%
\url{https://doi.org/10.1109/TMTT.1974.1128209}
\showDOI{\tempurl}


\bibitem[Bradbury et~al\mbox{.}(2018)]%
        {jax2018github}
\bibfield{author}{\bibinfo{person}{James Bradbury}, \bibinfo{person}{Roy
  Frostig}, \bibinfo{person}{Peter Hawkins}, \bibinfo{person}{Matthew~James
  Johnson}, \bibinfo{person}{Chris Leary}, \bibinfo{person}{Dougal Maclaurin},
  \bibinfo{person}{George Necula}, \bibinfo{person}{Adam Paszke},
  \bibinfo{person}{Jake Vander{P}las}, \bibinfo{person}{Skye
  Wanderman-{M}ilne}, {and} \bibinfo{person}{Qiao Zhang}.}
  \bibinfo{year}{2018}\natexlab{}.
\newblock \bibinfo{booktitle}{\emph{{JAX}: composable transformations of
  {P}ython+{N}um{P}y programs}}.
\newblock
\urldef\tempurl%
\url{http://github.com/google/jax}
\showURL{%
\tempurl}


\bibitem[Briggs et~al\mbox{.}(2000)]%
        {mgrid_tut}
\bibfield{author}{\bibinfo{person}{William~L. Briggs},
  \bibinfo{person}{Van~Emden Henson}, {and} \bibinfo{person}{Steve~F.
  McCormick}.} \bibinfo{year}{2000}\natexlab{}.
\newblock \bibinfo{booktitle}{\emph{A Multigrid Tutorial, Second Edition}
  (\bibinfo{edition}{second} ed.)}.
\newblock \bibinfo{publisher}{Society for Industrial and Applied Mathematics},
  \bibinfo{address}{Philadelphia, PA, USA}.
\newblock
\urldef\tempurl%
\url{https://doi.org/10.1137/1.9780898719505}
\showDOI{\tempurl}


\bibitem[Br\"{o}ker(2003)]%
        {BrokerThesis}
\bibfield{author}{\bibinfo{person}{Oliver Br\"{o}ker}.}
  \bibinfo{year}{2003}\natexlab{}.
\newblock \emph{\bibinfo{title}{Parallel multigrid methods using sparse
  approximate inverses}}.
\newblock \bibinfo{thesistype}{Ph.\,D. Dissertation}. \bibinfo{school}{ETH
  Zurich}.
\newblock
\urldef\tempurl%
\url{https://doi.org/10.3929/ETHZ-A-004617648}
\showDOI{\tempurl}


\bibitem[Brown et~al\mbox{.}(2021)]%
        {JBrown_etal_2019a}
\bibfield{author}{\bibinfo{person}{J. Brown}, \bibinfo{person}{Y. He},
  \bibinfo{person}{S. Mac{L}achlan}, \bibinfo{person}{M. Menickelly}, {and}
  \bibinfo{person}{S.M. Wild}.} \bibinfo{year}{2021}\natexlab{}.
\newblock \showarticletitle{Tuning multigrid methods with robust optimization}.
\newblock \bibinfo{journal}{\emph{SIAM J. Sci. Comput.}} \bibinfo{volume}{43},
  \bibinfo{number}{1} (\bibinfo{year}{2021}), \bibinfo{pages}{A109–A138}.
\newblock


\bibitem[Bruna et~al\mbox{.}(2013)]%
        {BrunaSpectral2013}
\bibfield{author}{\bibinfo{person}{Joan Bruna}, \bibinfo{person}{Wojciech
  Zaremba}, \bibinfo{person}{Arthur Szlam}, {and} \bibinfo{person}{Yann
  LeCun}.} \bibinfo{year}{2013}\natexlab{}.
\newblock \bibinfo{title}{Spectral Networks and Locally Connected Networks on
  Graphs}.
\newblock
\newblock
\urldef\tempurl%
\url{http://arxiv.org/abs/1312.6203}
\showURL{%
\tempurl}


\bibitem[Bröker and Grote(2002)]%
        {BrokerSPAIRelaxation2002}
\bibfield{author}{\bibinfo{person}{Oliver Bröker} {and}
  \bibinfo{person}{Marcus~J. Grote}.} \bibinfo{year}{2002}\natexlab{}.
\newblock \showarticletitle{Sparse approximate inverse smoothers for geometric
  and algebraic multigrid}.
\newblock \bibinfo{journal}{\emph{Applied Numerical Mathematics}}
  \bibinfo{volume}{41} (\bibinfo{date}{4} \bibinfo{year}{2002}),
  \bibinfo{pages}{61--80}.
\newblock
Issue 1.
\showISSN{0168-9274}
\urldef\tempurl%
\url{https://doi.org/10.1016/S0168-9274(01)00110-6}
\showDOI{\tempurl}


\bibitem[Chow and Patel(2015)]%
        {chow2015ilu}
\bibfield{author}{\bibinfo{person}{Edmond Chow} {and} \bibinfo{person}{Aftab
  Patel}.} \bibinfo{year}{2015}\natexlab{}.
\newblock \showarticletitle{Fine-Grained Parallel Incomplete {LU}
  Factorization}.
\newblock \bibinfo{journal}{\emph{SIAM Journal on Scientific Computing}}
  \bibinfo{volume}{37}, \bibinfo{number}{2} (\bibinfo{year}{2015}),
  \bibinfo{pages}{C169--C193}.
\newblock
\urldef\tempurl%
\url{https://doi.org/10.1137/140968896}
\showDOI{\tempurl}


\bibitem[Dalton et~al\mbox{.}(2015a)]%
        {2015_DaBaMeOlGa_merge}
\bibfield{author}{\bibinfo{person}{S. Dalton}, \bibinfo{person}{S. Baxter},
  \bibinfo{person}{D. Merrill}, \bibinfo{person}{L. Olson}, {and}
  \bibinfo{person}{M. Garland}.} \bibinfo{year}{2015}\natexlab{a}.
\newblock \showarticletitle{Optimizing Sparse Matrix Operations on {GPU}s Using
  Merge Path}. In \bibinfo{booktitle}{\emph{Parallel and Distributed Processing
  Symposium (IPDPS), 2015 IEEE International}}. \bibinfo{publisher}{IEEE},
  \bibinfo{address}{New York, NY, USA}, \bibinfo{pages}{407--416}.
\newblock
\showISSN{1530-2075}
\urldef\tempurl%
\url{https://doi.org/10.1109/IPDPS.2015.98}
\showDOI{\tempurl}


\bibitem[Dalton et~al\mbox{.}(2015b)]%
        {dalton2015spgemm}
\bibfield{author}{\bibinfo{person}{Steven Dalton}, \bibinfo{person}{Luke
  Olson}, {and} \bibinfo{person}{Nathan Bell}.}
  \bibinfo{year}{2015}\natexlab{b}.
\newblock \showarticletitle{Optimizing Sparse Matrix—Matrix Multiplication
  for the {GPU}}.
\newblock \bibinfo{journal}{\emph{ACM Trans. Math. Softw.}}
  \bibinfo{volume}{41}, \bibinfo{number}{4}, Article \bibinfo{articleno}{25}
  (\bibinfo{date}{oct} \bibinfo{year}{2015}), \bibinfo{numpages}{20}~pages.
\newblock
\showISSN{0098-3500}
\urldef\tempurl%
\url{https://doi.org/10.1145/2699470}
\showDOI{\tempurl}


\bibitem[Davis and Hu(2011)]%
        {suitesparse}
\bibfield{author}{\bibinfo{person}{Timothy~A. Davis} {and}
  \bibinfo{person}{Yifan Hu}.} \bibinfo{year}{2011}\natexlab{}.
\newblock \showarticletitle{The University of Florida Sparse Matrix
  Collection}.
\newblock \bibinfo{journal}{\emph{ACM Trans. Math. Software}}
  \bibinfo{volume}{38} (\bibinfo{date}{11} \bibinfo{year}{2011}),
  \bibinfo{pages}{25}.
\newblock
Issue 1.
\showISSN{00983500}
\urldef\tempurl%
\url{https://doi.org/10.1145/2049662.2049663}
\showDOI{\tempurl}


\bibitem[Demmel et~al\mbox{.}(1999)]%
        {Demmel1999}
\bibfield{author}{\bibinfo{person}{James~W. Demmel},
  \bibinfo{person}{Stanley~C. Eisenstat}, \bibinfo{person}{John~R. Gilbert},
  \bibinfo{person}{Xiaoye~S. Li}, {and} \bibinfo{person}{Joseph W.~H. Liu}.}
  \bibinfo{year}{1999}\natexlab{}.
\newblock \showarticletitle{A Supernodal Approach to Sparse Partial Pivoting}.
\newblock \bibinfo{journal}{\emph{SIAM J. Matrix Anal. Appl.}}
  \bibinfo{volume}{20} (\bibinfo{date}{1} \bibinfo{year}{1999}),
  \bibinfo{pages}{720--755}.
\newblock
Issue 3.
\showISSN{0895-4798}
\urldef\tempurl%
\url{https://doi.org/10.1137/S0895479895291765}
\showDOI{\tempurl}


\bibitem[Farrell et~al\mbox{.}(2021)]%
        {PFarrell_etal_2019a}
\bibfield{author}{\bibinfo{person}{P.~E. Farrell}, \bibinfo{person}{Y. He},
  {and} \bibinfo{person}{S. Mac{L}achlan}.} \bibinfo{year}{2021}\natexlab{}.
\newblock \showarticletitle{A local {F}ourier analysis of additive {V}anka
  relaxation for the {S}tokes equations}.
\newblock \bibinfo{journal}{\emph{Numer. Linear Alg. Appl.}}
  \bibinfo{volume}{28}, \bibinfo{number}{3} (\bibinfo{year}{2021}),
  \bibinfo{pages}{e2306}.
\newblock


\bibitem[Fey and Lenssen(2019)]%
        {pyg}
\bibfield{author}{\bibinfo{person}{Matthias Fey} {and}
  \bibinfo{person}{Jan~Eric Lenssen}.} \bibinfo{year}{2019}\natexlab{}.
\newblock \bibinfo{title}{Fast Graph Representation Learning with PyTorch
  Geometric}.
\newblock
\newblock
\urldef\tempurl%
\url{https://github.com/pyg-team/pytorch_geometric}
\showURL{%
\tempurl}


\bibitem[Gaihre et~al\mbox{.}(2022)]%
        {gaihre2022gsofa}
\bibfield{author}{\bibinfo{person}{Anil Gaihre}, \bibinfo{person}{Xiaoye~Sherry
  Li}, {and} \bibinfo{person}{Hang Liu}.} \bibinfo{year}{2022}\natexlab{}.
\newblock \showarticletitle{{GSoFa}: Scalable Sparse Symbolic {LU}
  Factorization on {GPUs}}.
\newblock \bibinfo{journal}{\emph{IEEE Transactions on Parallel and Distributed
  Systems}}  \bibinfo{volume}{33} (\bibinfo{date}{4} \bibinfo{year}{2022}),
  \bibinfo{pages}{1015--1026}.
\newblock
Issue 4.
\showISSN{15582183}
\urldef\tempurl%
\url{https://doi.org/10.1109/TPDS.2021.3090316}
\showDOI{\tempurl}


\bibitem[Gander and Kwok(2012)]%
        {gander2012robin}
\bibfield{author}{\bibinfo{person}{Martin~J. Gander} {and}
  \bibinfo{person}{Felix Kwok}.} \bibinfo{year}{2012}\natexlab{}.
\newblock \showarticletitle{Best {R}obin Parameters for Optimized {S}chwarz
  Methods at Cross Points}.
\newblock \bibinfo{journal}{\emph{SIAM Journal on Scientific Computing}}
  \bibinfo{volume}{34}, \bibinfo{number}{4} (\bibinfo{year}{2012}),
  \bibinfo{pages}{A1849--A1879}.
\newblock
\urldef\tempurl%
\url{https://doi.org/10.1137/110837218}
\showDOI{\tempurl}


\bibitem[Gebremedhin and Walther(2020)]%
        {Gebremedhin2020ADSurvey}
\bibfield{author}{\bibinfo{person}{Assefaw~H. Gebremedhin} {and}
  \bibinfo{person}{Andrea Walther}.} \bibinfo{year}{2020}\natexlab{}.
\newblock \showarticletitle{An introduction to algorithmic differentiation}.
\newblock \bibinfo{journal}{\emph{WIREs Data Mining and Knowledge Discovery}}
  \bibinfo{volume}{10} (\bibinfo{date}{1} \bibinfo{year}{2020}),
  \bibinfo{pages}{e1334}.
\newblock
Issue 1.
\showISSN{1942-4787}
\urldef\tempurl%
\url{https://doi.org/10.1002/widm.1334}
\showDOI{\tempurl}


\bibitem[George et~al\mbox{.}(1993)]%
        {george1993graph}
\bibfield{editor}{\bibinfo{person}{Alan George}, \bibinfo{person}{John~R.
  Gilbert}, {and} \bibinfo{person}{Joseph W.~H. Liu}} (Eds.).
  \bibinfo{year}{1993}\natexlab{}.
\newblock \bibinfo{booktitle}{\emph{Graph Theory and Sparse Matrix
  Computation}}.
\newblock \bibinfo{publisher}{Springer}, \bibinfo{address}{New York, NY ,USA}.
\newblock
\urldef\tempurl%
\url{https://doi.org/10.1007/978-1-4613-8369-7}
\showDOI{\tempurl}


\bibitem[Giles(2008)]%
        {MatAutodiff2008}
\bibfield{author}{\bibinfo{person}{Mike~B. Giles}.}
  \bibinfo{year}{2008}\natexlab{}.
\newblock \showarticletitle{Collected matrix derivative results for forward and
  reverse mode algorithmic differentiation}.
\newblock \bibinfo{journal}{\emph{Lecture Notes in Computational Science and
  Engineering}}  \bibinfo{volume}{64 LNCSE} (\bibinfo{year}{2008}),
  \bibinfo{pages}{35--44}.
\newblock
\showISBNx{9783540689355}
\showISSN{14397358}
\urldef\tempurl%
\url{https://doi.org/10.1007/978-3-540-68942-3_4/COVER}
\showDOI{\tempurl}


\bibitem[Greenfeld et~al\mbox{.}(2019)]%
        {GreenfeldInterpolation2019}
\bibfield{author}{\bibinfo{person}{Daniel Greenfeld}, \bibinfo{person}{Meirav
  Galun}, \bibinfo{person}{Ronen Basri}, \bibinfo{person}{Irad Yavneh}, {and}
  \bibinfo{person}{Ron Kimmel}.} \bibinfo{year}{2019}\natexlab{}.
\newblock \showarticletitle{Learning to Optimize Multigrid {PDE} Solvers}. In
  \bibinfo{booktitle}{\emph{Proceedings of the 36th International Conference on
  Machine Learning}} \emph{(\bibinfo{series}{Proceedings of Machine Learning
  Research}, Vol.~\bibinfo{volume}{97})},
  \bibfield{editor}{\bibinfo{person}{Kamalika Chaudhuri} {and}
  \bibinfo{person}{Ruslan Salakhutdinov}} (Eds.). \bibinfo{publisher}{PMLR},
  \bibinfo{pages}{2415--2423}.
\newblock
\urldef\tempurl%
\url{https://proceedings.mlr.press/v97/greenfeld19a.html}
\showURL{%
\tempurl}


\bibitem[Grote and Huckle(1997)]%
        {grotehuckle1997spai}
\bibfield{author}{\bibinfo{person}{Marcus~J. Grote} {and}
  \bibinfo{person}{Thomas Huckle}.} \bibinfo{year}{1997}\natexlab{}.
\newblock \showarticletitle{Parallel Preconditioning with Sparse Approximate
  Inverses}.
\newblock \bibinfo{journal}{\emph{SIAM Journal on Scientific Computing}}
  \bibinfo{volume}{18}, \bibinfo{number}{3} (\bibinfo{year}{1997}),
  \bibinfo{pages}{838--853}.
\newblock
\urldef\tempurl%
\url{https://doi.org/10.1137/S1064827594276552}
\showDOI{\tempurl}


\bibitem[Guo et~al\mbox{.}(2016)]%
        {2015_GuGrOl_gpu}
\bibfield{author}{\bibinfo{person}{Dahai Guo}, \bibinfo{person}{William Gropp},
  {and} \bibinfo{person}{Luke~N Olson}.} \bibinfo{year}{2016}\natexlab{}.
\newblock \showarticletitle{A hybrid format for better performance of sparse
  matrix-vector multiplication on a {GPU}}.
\newblock \bibinfo{journal}{\emph{The International Journal of High Performance
  Computing Applications}} \bibinfo{volume}{30}, \bibinfo{number}{1}
  (\bibinfo{year}{2016}), \bibinfo{pages}{103--120}.
\newblock
\urldef\tempurl%
\url{https://doi.org/10.1177/1094342015593156}
\showDOI{\tempurl}


\bibitem[Huang et~al\mbox{.}(2023)]%
        {Huang2021LearningRelaxation}
\bibfield{author}{\bibinfo{person}{Ru Huang}, \bibinfo{person}{Ruipeng Li},
  {and} \bibinfo{person}{Yuanzhe Xi}.} \bibinfo{year}{2023}\natexlab{}.
\newblock \showarticletitle{Learning Optimal Multigrid Smoothers via Neural
  Networks}.
\newblock \bibinfo{journal}{\emph{SIAM Journal on Scientific Computing}}
  \bibinfo{volume}{45}, \bibinfo{number}{3} (\bibinfo{year}{2023}),
  \bibinfo{pages}{S199--S225}.
\newblock
\urldef\tempurl%
\url{https://doi.org/10.1137/21M1430030}
\showDOI{\tempurl}


\bibitem[Huckle(2003)]%
        {HuckleFSAI2003}
\bibfield{author}{\bibinfo{person}{Thomas Huckle}.}
  \bibinfo{year}{2003}\natexlab{}.
\newblock \showarticletitle{Factorized Sparse Approximate Inverses for
  Preconditioning}.
\newblock \bibinfo{journal}{\emph{The Journal of Supercomputing}}
  \bibinfo{volume}{25} (\bibinfo{year}{2003}), \bibinfo{pages}{109--117}.
\newblock
\urldef\tempurl%
\url{https://doi.org/10.1023/A:1023988426844}
\showDOI{\tempurl}


\bibitem[Häusner et~al\mbox{.}(2023)]%
        {hausner2023neural}
\bibfield{author}{\bibinfo{person}{Paul Häusner}, \bibinfo{person}{Ozan
  Öktem}, {and} \bibinfo{person}{Jens Sjölund}.}
  \bibinfo{year}{2023}\natexlab{}.
\newblock \bibinfo{title}{Neural incomplete factorization: learning
  preconditioners for the conjugate gradient method}.
\newblock
\newblock
\showeprint[arxiv]{2305.16368}~[math.OC]


\bibitem[Kipf and Welling(2017)]%
        {kipf2017gcn}
\bibfield{author}{\bibinfo{person}{Thomas~N. Kipf} {and} \bibinfo{person}{Max
  Welling}.} \bibinfo{year}{2017}\natexlab{}.
\newblock \showarticletitle{Semi-Supervised Classification with Graph
  Convolutional Networks}. In \bibinfo{booktitle}{\emph{International
  Conference on Learning Representations}}.
\newblock


\bibitem[Knupp(2000)]%
        {knupp2000meshopt}
\bibfield{author}{\bibinfo{person}{Patrick~M. Knupp}.}
  \bibinfo{year}{2000}\natexlab{}.
\newblock \showarticletitle{Achieving finite element mesh quality via
  optimization of the {J}acobian matrix norm and associated quantities. {P}art
  {I}—a framework for surface mesh optimization}.
\newblock \bibinfo{journal}{\emph{Internat. J. Numer. Methods Engrg.}}
  \bibinfo{volume}{48}, \bibinfo{number}{3} (\bibinfo{year}{2000}),
  \bibinfo{pages}{401--420}.
\newblock
\urldef\tempurl%
\url{https://doi.org/10.1002/(SICI)1097-0207(20000530)48:3<401::AID-NME880>3.0.CO;2-D}
\showDOI{\tempurl}


\bibitem[Kolotilina and Yeremin(1993)]%
        {KolotilinaFSAITheory1993}
\bibfield{author}{\bibinfo{person}{L.~{Yu.} Kolotilina} {and}
  \bibinfo{person}{A.~{Yu.} Yeremin}.} \bibinfo{year}{1993}\natexlab{}.
\newblock \showarticletitle{Factorized Sparse Approximate Inverse
  Preconditionings {I}. {T}heory}.
\newblock \bibinfo{journal}{\emph{SIAM J. Matrix Anal. Appl.}}
  \bibinfo{volume}{14} (\bibinfo{date}{1} \bibinfo{year}{1993}),
  \bibinfo{pages}{45--58}.
\newblock
Issue 1.
\showISSN{0895-4798}
\urldef\tempurl%
\url{https://doi.org/10.1137/0614004}
\showDOI{\tempurl}


\bibitem[Kumar et~al\mbox{.}(2019)]%
        {rodrigolfa2019}
\bibfield{author}{\bibinfo{person}{Prashant Kumar}, \bibinfo{person}{Carmen
  Rodrigo}, \bibinfo{person}{Francisco~J. Gaspar}, {and}
  \bibinfo{person}{Cornelis~W. Oosterlee}.} \bibinfo{year}{2019}\natexlab{}.
\newblock \showarticletitle{On Local {F}ourier Analysis of Multigrid Methods
  for {PDEs} with Jumping and Random Coefficients}.
\newblock \bibinfo{journal}{\emph{SIAM Journal on Scientific Computing}}
  \bibinfo{volume}{41} (\bibinfo{date}{1} \bibinfo{year}{2019}),
  \bibinfo{pages}{A1385--A1413}.
\newblock
Issue 3.
\showISSN{1064-8275}
\urldef\tempurl%
\url{https://doi.org/10.1137/18M1173769}
\showDOI{\tempurl}


\bibitem[Nolan(1953)]%
        {NolanAutodiff1953}
\bibfield{author}{\bibinfo{person}{John~F. Nolan}.}
  \bibinfo{year}{1953}\natexlab{}.
\newblock \emph{\bibinfo{title}{Analytical differentiation on a digital
  computer}}.
\newblock \bibinfo{thesistype}{Ph.\,D. Dissertation}.
  \bibinfo{school}{Massachusetts Institute of Technology}.
\newblock
\urldef\tempurl%
\url{https://dspace.mit.edu/handle/1721.1/12297}
\showURL{%
\tempurl}


\bibitem[Oosterlee and Wienands(2003)]%
        {GeneticMultigrid2003}
\bibfield{author}{\bibinfo{person}{C.~W. Oosterlee} {and} \bibinfo{person}{R.
  Wienands}.} \bibinfo{year}{2003}\natexlab{}.
\newblock \showarticletitle{A genetic search for optimal multigrid components
  within a {F}ourier analysis setting}.
\newblock \bibinfo{journal}{\emph{SIAM Journal on Scientific Computing}}
  \bibinfo{volume}{24} (\bibinfo{year}{2003}), \bibinfo{pages}{924--944}.
\newblock
Issue 3.
\showISSN{10648275}
\urldef\tempurl%
\url{https://doi.org/10.1137/S1064827501397950}
\showDOI{\tempurl}


\bibitem[Paszke et~al\mbox{.}(2019)]%
        {pytorch}
\bibfield{author}{\bibinfo{person}{Adam Paszke}, \bibinfo{person}{Sam Gross},
  \bibinfo{person}{Francisco Massa}, \bibinfo{person}{Adam Lerer},
  \bibinfo{person}{James Bradbury}, \bibinfo{person}{Gregory Chanan},
  \bibinfo{person}{Trevor Killeen}, \bibinfo{person}{Zeming Lin},
  \bibinfo{person}{Natalia Gimelshein}, \bibinfo{person}{Luca Antiga},
  \bibinfo{person}{Alban Desmaison}, \bibinfo{person}{Andreas K{\"{o}}pf},
  \bibinfo{person}{Edward~Z. Yang}, \bibinfo{person}{Zach DeVito},
  \bibinfo{person}{Martin Raison}, \bibinfo{person}{Alykhan Tejani},
  \bibinfo{person}{Sasank Chilamkurthy}, \bibinfo{person}{Benoit Steiner},
  \bibinfo{person}{Lu Fang}, \bibinfo{person}{Junjie Bai}, {and}
  \bibinfo{person}{Soumith Chintala}.} \bibinfo{year}{2019}\natexlab{}.
\newblock \showarticletitle{{PyTorch}: An Imperative Style, High-Performance
  Deep Learning Library}.
\newblock \bibinfo{journal}{\emph{CoRR}}  \bibinfo{volume}{abs/1912.01703}
  (\bibinfo{year}{2019}).
\newblock


\bibitem[Peng and Tan(2020)]%
        {peng2019slu}
\bibfield{author}{\bibinfo{person}{Shaoyi Peng} {and} \bibinfo{person}{Sheldon
  X.-D. Tan}.} \bibinfo{year}{2020}\natexlab{}.
\newblock \showarticletitle{{GLU3.0}: Fast {GPU}-based Parallel Sparse {LU}
  Factorization for Circuit Simulation}.
\newblock \bibinfo{journal}{\emph{IEEE Design \& Test}} \bibinfo{volume}{37},
  \bibinfo{number}{3} (\bibinfo{year}{2020}), \bibinfo{pages}{78--90}.
\newblock
\urldef\tempurl%
\url{https://doi.org/10.1109/MDAT.2020.2974910}
\showDOI{\tempurl}


\bibitem[Polyak(1964)]%
        {polyak1964heavyball}
\bibfield{author}{\bibinfo{person}{B.T. Polyak}.}
  \bibinfo{year}{1964}\natexlab{}.
\newblock \showarticletitle{Some methods of speeding up the convergence of
  iteration methods}.
\newblock \bibinfo{journal}{\emph{U. S. S. R. Comput. Math. and Math. Phys.}}
  \bibinfo{volume}{4}, \bibinfo{number}{5} (\bibinfo{year}{1964}),
  \bibinfo{pages}{1--17}.
\newblock
\showISSN{0041-5553}
\urldef\tempurl%
\url{https://doi.org/10.1016/0041-5553(64)90137-5}
\showDOI{\tempurl}


\bibitem[Quarteroni et~al\mbox{.}(2006)]%
        {quarteroni2007numerical}
\bibfield{author}{\bibinfo{person}{Alfio Quarteroni}, \bibinfo{person}{Riccardo
  Sacco}, {and} \bibinfo{person}{Fausto Saleri}.}
  \bibinfo{year}{2006}\natexlab{}.
\newblock \bibinfo{booktitle}{\emph{Numerical Mathematics}
  (\bibinfo{edition}{2} ed.)}.
\newblock \bibinfo{publisher}{Springer}, \bibinfo{address}{Berlin, Germany}.
\newblock


\bibitem[Saad(2003)]%
        {saad2003iterative}
\bibfield{author}{\bibinfo{person}{Yousef Saad}.}
  \bibinfo{year}{2003}\natexlab{}.
\newblock \bibinfo{booktitle}{\emph{{Iterative Methods for Sparse Linear
  Systems}} (\bibinfo{edition}{second} ed.)}.
\newblock \bibinfo{publisher}{SIAM}, \bibinfo{address}{Philadelphia, PA, USA}.
\newblock
\showISBNx{978-0-89871-534-7}
\urldef\tempurl%
\url{https://doi.org/10.1137/1.9780898718003}
\showDOI{\tempurl}


\bibitem[Su et~al\mbox{.}(2020)]%
        {Capellini2020}
\bibfield{author}{\bibinfo{person}{Jiya Su}, \bibinfo{person}{Feng Zhang},
  \bibinfo{person}{Weifeng Liu}, \bibinfo{person}{Bingsheng He},
  \bibinfo{person}{Ruofan Wu}, \bibinfo{person}{Xiaoyong Du}, {and}
  \bibinfo{person}{Rujia Wang}.} \bibinfo{year}{2020}\natexlab{}.
\newblock \showarticletitle{{CapelliniSpTRSV}: A Thread-Level
  Synchronization-Free Sparse Triangular Solve on {GPU}s}. In
  \bibinfo{booktitle}{\emph{Proceedings of the 49th International Conference on
  Parallel Processing}} (Edmonton, AB, Canada) \emph{(\bibinfo{series}{ICPP
  '20})}. \bibinfo{publisher}{Association for Computing Machinery},
  \bibinfo{address}{New York, NY, USA}, Article \bibinfo{articleno}{2},
  \bibinfo{numpages}{11}~pages.
\newblock
\showISBNx{9781450388160}
\urldef\tempurl%
\url{https://doi.org/10.1145/3404397.3404400}
\showDOI{\tempurl}


\bibitem[Taghibakhshi et~al\mbox{.}(2022)]%
        {taghibakhshi2022mloras}
\bibfield{author}{\bibinfo{person}{Ali Taghibakhshi}, \bibinfo{person}{Nicolas
  Nytko}, \bibinfo{person}{Tareq Zaman}, \bibinfo{person}{Scott MacLachlan},
  \bibinfo{person}{Luke Olson}, {and} \bibinfo{person}{Matthew West}.}
  \bibinfo{year}{2022}\natexlab{}.
\newblock \showarticletitle{Learning Interface Conditions in Domain
  Decomposition Solvers}. In \bibinfo{booktitle}{\emph{Advances in Neural
  Information Processing Systems}},
  \bibfield{editor}{\bibinfo{person}{S.~Koyejo}, \bibinfo{person}{S.~Mohamed},
  \bibinfo{person}{A.~Agarwal}, \bibinfo{person}{D.~Belgrave},
  \bibinfo{person}{K.~Cho}, {and} \bibinfo{person}{A.~Oh}} (Eds.),
  Vol.~\bibinfo{volume}{35}. \bibinfo{publisher}{Curran Associates, Inc.},
  \bibinfo{pages}{7222--7235}.
\newblock


\bibitem[Tao et~al\mbox{.}(2014)]%
        {SpTMv2014}
\bibfield{author}{\bibinfo{person}{Yuan Tao}, \bibinfo{person}{Yangdong Deng},
  \bibinfo{person}{Shuai Mu}, \bibinfo{person}{Mingfa Zhu},
  \bibinfo{person}{Limin Xiao}, \bibinfo{person}{Li Ruan}, {and}
  \bibinfo{person}{Zhibin Huang}.} \bibinfo{year}{2014}\natexlab{}.
\newblock \showarticletitle{Atomic reduction based sparse matrix-transpose
  vector multiplication on {GPUs}}.
\newblock \bibinfo{journal}{\emph{Proceedings of the International Conference
  on Parallel and Distributed Systems - ICPADS}}  \bibinfo{volume}{2015-April}
  (\bibinfo{year}{2014}), \bibinfo{pages}{987--992}.
\newblock
\showISBNx{9781479976157}
\showISSN{15219097}
\urldef\tempurl%
\url{https://doi.org/10.1109/PADSW.2014.7097920}
\showDOI{\tempurl}


\bibitem[Thompson et~al\mbox{.}(2023)]%
        {thompsonlfa2021}
\bibfield{author}{\bibinfo{person}{Jeremy~L. Thompson}, \bibinfo{person}{Jed
  Brown}, {and} \bibinfo{person}{Yunhui He}.} \bibinfo{year}{2023}\natexlab{}.
\newblock \showarticletitle{Local Fourier Analysis of p-Multigrid for
  High-Order Finite Element Operators}.
\newblock \bibinfo{journal}{\emph{SIAM Journal on Scientific Computing}}
  \bibinfo{volume}{45}, \bibinfo{number}{3} (\bibinfo{year}{2023}),
  \bibinfo{pages}{S351--S370}.
\newblock
\urldef\tempurl%
\url{https://doi.org/10.1137/21M1431199}
\showDOI{\tempurl}


\bibitem[Toselli and Widlund(2005)]%
        {Toselli2005}
\bibfield{author}{\bibinfo{person}{Andrea Toselli} {and}
  \bibinfo{person}{Olof~B. Widlund}.} \bibinfo{year}{2005}\natexlab{}.
\newblock \bibinfo{booktitle}{\emph{Domain Decomposition Methods {\textemdash}
  Algorithms and Theory}}.
\newblock \bibinfo{publisher}{Springer}, \bibinfo{address}{Berlin Heidelberg}.
\newblock
\urldef\tempurl%
\url{https://doi.org/10.1007/b137868}
\showDOI{\tempurl}


\bibitem[Wang et~al\mbox{.}(2020)]%
        {dgl}
\bibfield{author}{\bibinfo{person}{Minjie Wang}, \bibinfo{person}{Da Zheng},
  \bibinfo{person}{Zihao Ye}, \bibinfo{person}{Quan Gan},
  \bibinfo{person}{Mufei Li}, \bibinfo{person}{Xiang Song},
  \bibinfo{person}{Jinjing Zhou}, \bibinfo{person}{Chao Ma},
  \bibinfo{person}{Lingfan Yu}, \bibinfo{person}{Yu Gai},
  \bibinfo{person}{Tianjun Xiao}, \bibinfo{person}{Tong He},
  \bibinfo{person}{George Karypis}, \bibinfo{person}{Jinyang Li}, {and}
  \bibinfo{person}{Zheng Zhang}.} \bibinfo{year}{2020}\natexlab{}.
\newblock \bibinfo{title}{Deep Graph Library: A Graph-Centric,
  Highly-Performant Package for Graph Neural Networks}.
\newblock
\newblock
\showeprint[arxiv]{1909.01315}~[cs.LG]


\bibitem[Wang et~al\mbox{.}(2021)]%
        {WangCUDASPAI2021}
\bibfield{author}{\bibinfo{person}{Yizhou Wang}, \bibinfo{person}{Wenhao Li},
  {and} \bibinfo{person}{Jiaquan Gao}.} \bibinfo{year}{2021}\natexlab{}.
\newblock \showarticletitle{A parallel sparse approximate inverse
  preconditioning algorithm based on {MPI} and {CUDA}}.
\newblock \bibinfo{journal}{\emph{BenchCouncil Transactions on Benchmarks,
  Standards and Evaluations}}  \bibinfo{volume}{1} (\bibinfo{date}{10}
  \bibinfo{year}{2021}), \bibinfo{pages}{100007}.
\newblock
Issue 1.
\showISSN{2772-4859}
\urldef\tempurl%
\url{https://doi.org/10.1016/J.TBENCH.2021.100007}
\showDOI{\tempurl}


\bibitem[Wienands and Joppich(2004)]%
        {wienandspfa2004}
\bibfield{author}{\bibinfo{person}{Roman Wienands} {and}
  \bibinfo{person}{Wolfgang Joppich}.} \bibinfo{year}{2004}\natexlab{}.
\newblock \bibinfo{booktitle}{\emph{Practical Fourier Analysis for Multigrid
  Methods}}.
\newblock \bibinfo{publisher}{Chapman and Hall/CRC}, \bibinfo{address}{New
  York, NY, USA}.
\newblock


\bibitem[Wu et~al\mbox{.}(2021)]%
        {Wu2021GraphnetSurvey}
\bibfield{author}{\bibinfo{person}{Zonghan Wu}, \bibinfo{person}{Shirui Pan},
  \bibinfo{person}{Fengwen Chen}, \bibinfo{person}{Guodong Long},
  \bibinfo{person}{Chengqi Zhang}, {and} \bibinfo{person}{Philip~S. Yu}.}
  \bibinfo{year}{2021}\natexlab{}.
\newblock \showarticletitle{A Comprehensive Survey on Graph Neural Networks}.
\newblock \bibinfo{journal}{\emph{IEEE Transactions on Neural Networks and
  Learning Systems}}  \bibinfo{volume}{32} (\bibinfo{date}{1}
  \bibinfo{year}{2021}), \bibinfo{pages}{4--24}.
\newblock
Issue 1.
\showISSN{2162-237X}
\urldef\tempurl%
\url{https://doi.org/10.1109/TNNLS.2020.2978386}
\showDOI{\tempurl}


\bibitem[Yang et~al\mbox{.}(2016)]%
        {CiteSeerDataset}
\bibfield{author}{\bibinfo{person}{Zhilin Yang}, \bibinfo{person}{William
  Cohen}, {and} \bibinfo{person}{Ruslan Salakhudinov}.}
  \bibinfo{year}{2016}\natexlab{}.
\newblock \showarticletitle{Revisiting Semi-Supervised Learning with Graph
  Embeddings}. In \bibinfo{booktitle}{\emph{Proceedings of The 33rd
  International Conference on Machine Learning}}
  \emph{(\bibinfo{series}{Proceedings of Machine Learning Research},
  Vol.~\bibinfo{volume}{48})}, \bibfield{editor}{\bibinfo{person}{Maria~Florina
  Balcan} {and} \bibinfo{person}{Kilian~Q. Weinberger}} (Eds.).
  \bibinfo{publisher}{PMLR}, \bibinfo{address}{New York, New York, USA},
  \bibinfo{pages}{40--48}.
\newblock
\urldef\tempurl%
\url{https://proceedings.mlr.press/v48/yanga16.html}
\showURL{%
\tempurl}


\bibitem[Zeng et~al\mbox{.}(2020)]%
        {FlickrDataset}
\bibfield{author}{\bibinfo{person}{Hanqing Zeng}, \bibinfo{person}{Hongkuan
  Zhou}, \bibinfo{person}{Ajitesh Srivastava}, \bibinfo{person}{Rajgopal
  Kannan}, {and} \bibinfo{person}{Viktor Prasanna}.}
  \bibinfo{year}{2020}\natexlab{}.
\newblock \showarticletitle{{GraphSAINT}: Graph Sampling Based Inductive
  Learning Method}. In \bibinfo{booktitle}{\emph{International Conference on
  Learning Representations}}.
\newblock
\urldef\tempurl%
\url{https://openreview.net/forum?id=BJe8pkHFwS}
\showURL{%
\tempurl}


\bibitem[Zhao et~al\mbox{.}(2021)]%
        {zhao2021sflu}
\bibfield{author}{\bibinfo{person}{Jianqi Zhao}, \bibinfo{person}{Yao Wen},
  \bibinfo{person}{Yuchen Luo}, \bibinfo{person}{Zhou Jin},
  \bibinfo{person}{Weifeng Liu}, {and} \bibinfo{person}{Zhenya Zhou}.}
  \bibinfo{year}{2021}\natexlab{}.
\newblock \showarticletitle{{SFLU}: Synchronization-Free Sparse {LU}
  Factorization for Fast Circuit Simulation on {GPUs}}. In
  \bibinfo{booktitle}{\emph{2021 58th ACM/IEEE Design Automation Conference
  (DAC)}}. \bibinfo{pages}{37--42}.
\newblock
\urldef\tempurl%
\url{https://doi.org/10.1109/DAC18074.2021.9586141}
\showDOI{\tempurl}


\end{thebibliography}
\end{document}